\documentclass[journal]{IEEEtran}
\usepackage{amsmath,amsfonts}
\usepackage{algorithmic}
\usepackage{algorithm}
\usepackage{array}   
\usepackage{booktabs, amsmath, bm}
\usepackage{wasysym}
\usepackage{multirow}
\usepackage[table]{xcolor}
\usepackage[caption=false,font=normalsize,labelfont=sf, labelformat=simple,textfont=sf]{subfig}

\usepackage[
    linkcolor=blue,    
    urlcolor=blue,     
    citecolor=blue,    
    colorlinks=true    
]{hyperref}
\usepackage{textcomp}
\usepackage{stfloats}
\usepackage{url}
\usepackage{verbatim}
\usepackage{graphicx}
\usepackage[nocompress]{cite}
\usepackage{subfiles}
\usepackage{amssymb}
\begin{document}

\title{EventFace: Event-Based Face Recognition via Structure-Driven Spatiotemporal Modeling}
 
\author{Qingguo Meng, Xingbo Dong, ~\IEEEmembership{Member,~IEEE}, Zhe Jin, ~\IEEEmembership{Member,~IEEE}, Massimo Tistarelli, ~\IEEEmembership{Senior Member,~IEEE}
\thanks{This work was supported by the National Natural Science Foundation of China under Grant 62376003. (\textit{Corresponding authors: Xingbo Dong, Zhe Jin})}
\thanks{Qingguo Meng, Xingbo Dong, Zhe Jin are with State Key Laboratory of Opto-Electronic Information Acquisition and Protection Technology, the Anhui Provincial Key Laboratory of Secure Artificial Intelligence, Anhui Provincial International Joint Research Center for Advanced Technology in Medical Imaging, and the School of Artificial Intelligence, Anhui University, Hefei, China (e-mail: mqg1024@163.com, xingbo.dong@ahu.edu.cn, jinzhe@ahu.edu.cn).}
\thanks{Massimo Tistarelli is with the Computer Vision Laboratory, University of
Sassari, 07100 Sassari, Italy (e-mail: tista@uniss.it).}}

\markboth{Journal of \LaTeX\ Class Files,~Vol.~14, No.~8, August~2021}%
{Shell \MakeLowercase{\textit{et al.}}: A Sample Article Using IEEEtran.cls for IEEE Journals}

\IEEEpubid{0000--0000/00\$00.00~\copyright~2021 IEEE} 

\maketitle

\begin{abstract}
    Event cameras offer a promising sensing modality for face recognition due to their inherent advantages in illumination robustness and privacy-friendliness. However, because event streams lack the stable photometric appearance relied upon by conventional RGB-based face recognition systems, we argue that event-based face recognition should model structure-driven spatiotemporal identity representations shaped by rigid facial motion and individual facial geometry.
Since dedicated datasets for event-based face recognition remain lacking, we construct EFace, a small-scale event-based face dataset captured under rigid facial motion. To learn effectively from this limited event data, we further propose EventFace, a framework for event-based face recognition that integrates spatial structure and temporal dynamics for identity modeling.
Specifically, we employ Low-Rank Adaptation (LoRA) to transfer structural facial priors from pretrained RGB face models to the event domain, thereby establishing a reliable spatial basis for identity modeling. Building on this foundation, we further introduce a Motion Prompt Encoder (MPE) to explicitly encode temporal features and a Spatiotemporal Modulator (STM) to fuse them with spatial features, thereby enhancing the representation of identity-relevant event patterns.
Extensive experiments demonstrate that EventFace achieves the best performance among the evaluated baselines, with a Rank-1 identification rate of 94.19\% and an equal error rate (EER) of 5.35\%.
Results further indicate that EventFace exhibits stronger robustness under degraded illumination than the competing methods. In addition, the learned representations exhibit reduced template reconstructability.
\end{abstract}

\begin{IEEEkeywords}
Face recognition, event-based vision, spatio-temporal modeling.
\end{IEEEkeywords}

\section{Introduction}
    \label{sec:intro}
    \IEEEPARstart{E}{vent} cameras adopt a sensing mechanism fundamentally different from conventional frame-based cameras. Instead of recording dense image frames at fixed intervals, they asynchronously capture pixel-level intensity changes when brightness variations exceed a contrast threshold~\cite{lichtsteiner2008128,brandli2014240}. This imaging principle endows them with high dynamic range, enabling robust sensing under challenging illumination conditions~\cite{gallego2020event,ren2026eventvggt,zhu2025depth,wang2026person}. Meanwhile, because event streams do not directly preserve dense texture information, they expose far less visually interpretable appearance content than RGB imagery, suggesting that event cameras may offer a more privacy-friendly sensing modality~\cite{gallego2020event,wang2026person,dong2023bullying10k}.

\begin{figure}[!t]
  \centering
  \captionsetup[subfloat]{font=footnotesize}
  \subfloat[Face recognition based on RGB frames and its limitations\label{fig:teaser-rgb}]{
    \includegraphics[width=\linewidth]{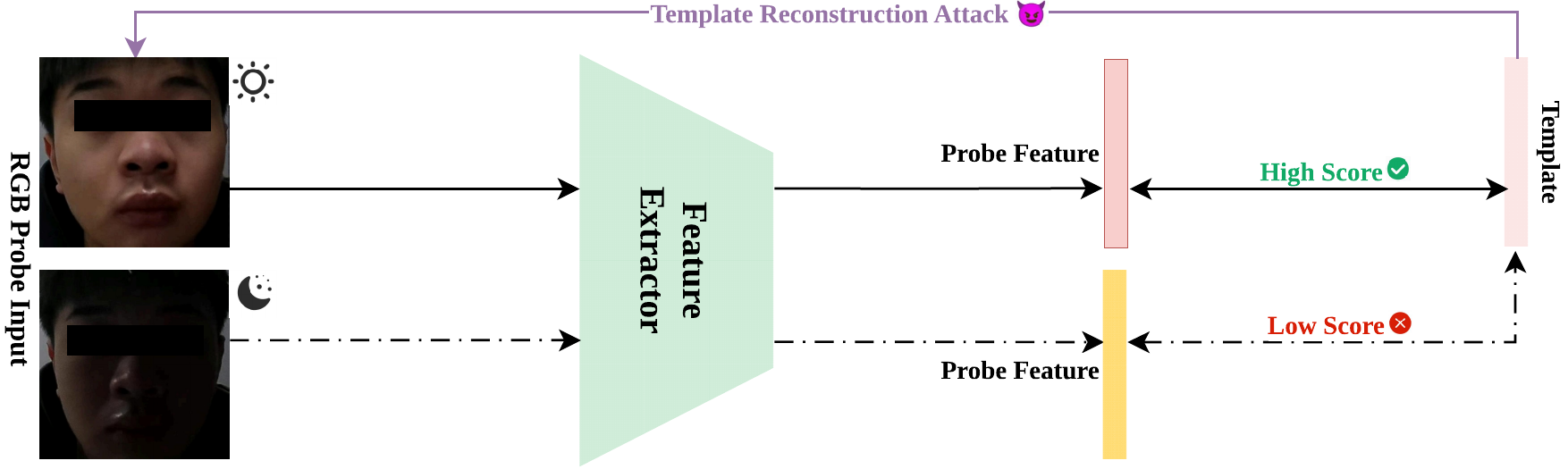}
  }

  \subfloat[Face recognition based on event data (ours)\label{fig:teaser-event}]{
    \includegraphics[width=\linewidth]{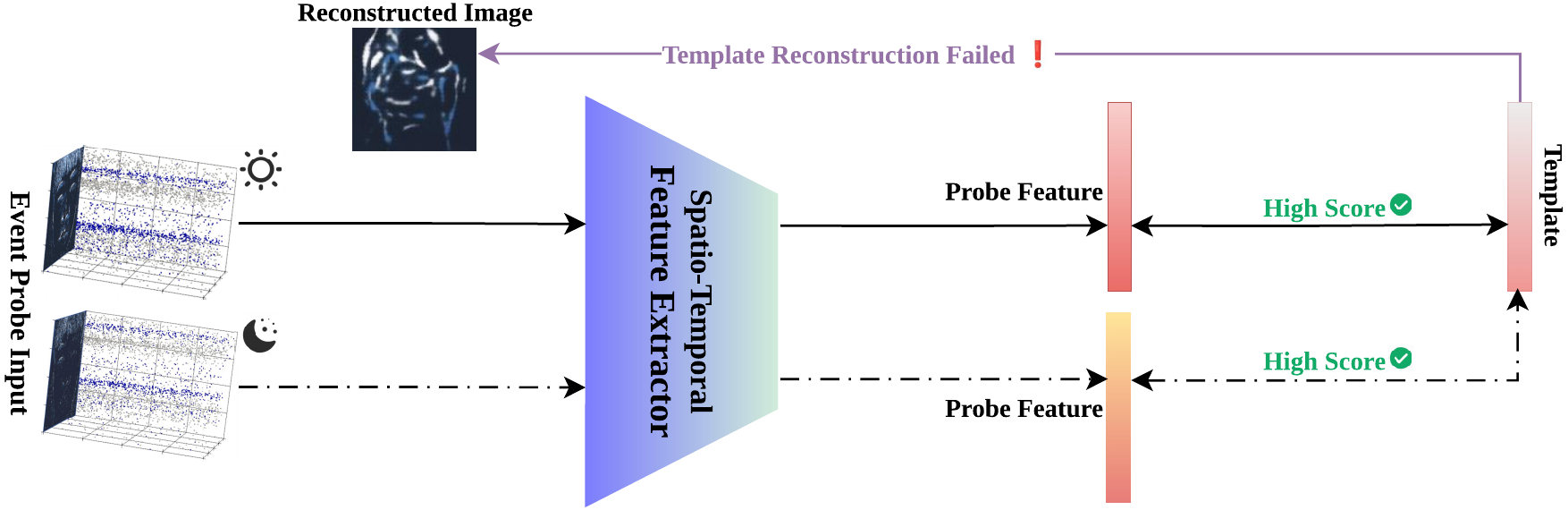}
  }
  \caption{(a) Traditional RGB face recognition is sensitive to lighting variations and vulnerable to template reconstruction attacks. (b) By contrast, event-based face recognition leverages structure-driven spatiotemporal patterns generated by rigid facial motion, achieving more stable recognition under varying illumination while reducing reconstructability from leaked templates.}
  \label{fig:teaser}
\end{figure}

These properties make event cameras a promising sensing modality for face recognition, as shown in Fig.~\ref{fig:teaser-event}. However, these sensing characteristics also fundamentally change how facial identity should be represented. In conventional face recognition~\cite{wang2022survey,guo2019survey,schroff2015facenet,deng2019arcface,kim2022adaface}, discriminative identity cues are largely conveyed by stable photometric appearance, such as texture and detailed facial patterns. As illustrated in Fig.~\ref{fig:teaser-rgb}, these methods are highly sensitive to lighting variations~\cite{shepley2019deep,baek2025low} and vulnerable to template reconstruction attacks~\cite{sun2025ensuring,kansy2023controllable,duong2020vec2face}.
\IEEEpubidadjcol
In contrast, event streams are triggered mainly by pixel-level intensity changes and therefore do not preserve such static appearance information. As a result, identity learning in the event domain cannot simply rely on appearance-based encoding. Instead, richer event responses often arise along facial contours and component boundaries during motion. Identity cues may thus emerge from the interaction between intrinsic facial structure and the temporal responses induced by rigid facial motion. This suggests that event-based face recognition may be better formulated as learning structure-driven spatiotemporal identity representations rather than relying primarily on appearance matching.

Despite this conceptual promise, realizing event-based face recognition in practice remains challenging. First, dedicated event-based face datasets remain lacking, while face recognition models typically require large-scale supervision to learn identity-discriminative representations that generalize to unseen subjects. Second, event data differs fundamentally from RGB imagery: it is sparse, asynchronous, and lacks dense appearance information. These characteristics prevent the direct use of conventional RGB face recognition pipelines and make it difficult to learn robust identity representations solely from event-domain supervision.

A practical way to address these challenges is to transfer prior knowledge from the RGB domain. Although RGB frames and event streams differ substantially in sensing mechanism, they are both generated from the same underlying facial structure~\cite{becattini2024neuromorphic}. Consequently, face recognition models pretrained on large-scale RGB datasets contain rich structural priors that can potentially benefit event-based recognition~\cite{kim2022adaface}. Motivated by this observation, we propose EventFace, a cross-modal adaptation framework that transfers facial structural priors from RGB face models to the event domain, while explicitly modeling the temporal dynamics unique to event streams.

Our framework adopts a two-stage learning strategy. First, structural priors from a pretrained RGB face recognition backbone are transferred to the event domain via Low-Rank Adaptation (LoRA)~\cite{hu2022lora}. This stage not only equips the model with transferable facial structural priors, but also alleviates its dependence on large-scale dedicated event-based face datasets. The adapted backbone is then frozen to better retain these transferred priors. Second, a Motion Prompt Encoder (MPE) is attached to each backbone stage to capture motion dynamics from event sequences at different semantic levels. The resulting spatial and temporal features are fused by a Spatiotemporal Modulator (STM), which enables mutual refinement between facial structure and motion features. Repeated across stages, this design progressively enhances spatiotemporal representations and yields robust identity features for event-based face recognition.

To provide the necessary event-domain supervision for structural prior transfer and temporal modeling, we construct EFace, a small-scale event-based rigid facial motion dataset with disjoint identities for training and testing. Moreover, a subset of the test set is collected under challenging illumination conditions.

In summary, this paper makes the following contributions:
\begin{itemize}
    \item We propose EventFace, a framework for event-based face recognition that transfers RGB facial structural priors and integrates motion dynamics for identity modeling.
    \item We introduce Structure-Aware Spatial Adaptation, a LoRA-based stage for RGB-to-event prior transfer from a pretrained RGB face recognition model. Building on this adapted backbone, we further introduce the MPE and the STM to capture motion-induced dynamics and progressively fuse them with spatial features across network depths.
    \item We construct EFace, a small-scale event-based face dataset captured under rigid facial motion, with disjoint training and testing identities and an illumination-degraded test subset for robustness evaluation.
    \item We conduct systematic evaluations of the proposed method in terms of identity recognition, illumination robustness, and template reconstruction resistance, providing comprehensive evidence for the practical potential of event cameras in face recognition.
\end{itemize}

\section{Related Work}
    \subsection{Appearance-Based Face Recognition}
Appearance-based face recognition has achieved remarkable progress over the past decade, driven by advances in backbone architectures, large-scale data collection, and discriminative learning strategies~\cite{wang2022survey,wang2021deep,guo2019survey}. Early deep face recognition methods predominantly relied on convolutional neural networks to learn hierarchical facial representations from static images~\cite{sun2014deep,taigman2014deepface,schroff2015facenet}. More recently, transformer-based backbones have also been introduced into face recognition by adapting vision transformers with identity-aware training strategies~\cite{dan2023transface,dan2025transface++}. In parallel, the availability of large-scale face datasets, often containing millions of images from a large number of identities, has been equally important, as it provides the broad intra-class and inter-class variation required for learning highly discriminative identity representations~\cite{guo2016ms,cao2018vggface2,kemelmacher2016megaface}.

Beyond architectural and data-driven progress, the evolution of discriminative learning objectives has played a critical role in optimizing the feature embedding space. Early models formulated face recognition as a closed-set multi-class classification problem using standard Softmax~\cite{taigman2014deepface,sun2014deep}, which lacked explicit optimization for open-set verification. To address this, subsequent research explored metric learning paradigms~\cite{sun2014deepid2,schroff2015facenet} to directly optimize similarity distances, though they often suffered from combinatorial explosion during pair mining. Consequently, the field converged on margin-based classification objectives (e.g., SphereFace~\cite{liu2017sphereface}, CosFace~\cite{wang2018cosface}, ArcFace~\cite{deng2019arcface}, and AdaFace~\cite{kim2022adaface}). By elegantly introducing angular margins into the classification loss, these methods enforce intra-class compactness and inter-class discrepancy without the need for complex sampling strategies. Together, these advances have established a mature framework for appearance-based face recognition.

Despite their strong performance under standard conditions, deep learning–based appearance-based face recognition models often degrade significantly under low-light scenarios. Trained primarily on well-illuminated images, these models rely on photometric appearance cues that become unreliable when illumination is insufficient, leading to unstable identity embeddings~\cite{shepley2019deep}. Existing solutions typically incorporate image enhancement or restoration networks as preprocessing steps, which increase system complexity and may introduce artifacts or distortions that adversely affect identity-discriminative features~\cite{fan2024low,baek2025low}.

Beyond robustness, privacy and security have emerged as critical concerns in appearance-based face recognition~\cite{sun2025ensuring}. Recent studies demonstrate that identity embeddings learned from deep models may still retain sufficient information to enable face reconstruction or attribute inference, posing potential privacy risks~\cite{mai2018reconstruction,duong2020vec2face,kansy2023controllable}. To mitigate such threats, template protection strategies have been explored from two main perspectives. Cryptography-based approaches protect face templates through secure transformation or encryption, offering strong theoretical security guarantees but often requiring complex system design and additional computational overhead~\cite{kim2025idface,osorio2021stable,bauspiess2023hebi}. Alternatively, visual transformation–based methods apply perturbations, obfuscation, or adversarial noise to facial images or features to suppress identifiable appearance information, though they may compromise recognition performance or remain vulnerable to reconstruction attacks~\cite{mi2024privacy,yuan2022pro1,yuan2024pro2,wang2023privacy}. These limitations motivate the exploration of alternative sensing paradigms that can inherently reduce reliance on detailed facial appearance while maintaining discriminative identity representations.

\subsection{Event-Based Face Analysis}
Event cameras offer several advantages, including asynchronous sensing, high temporal resolution, and high dynamic range, which have motivated extensive research across a wide range of computer vision tasks~\cite{chakravarthi2024recent,gallego2020event}. However, compared with general vision problems, the application of event-based vision to face-related tasks remains relatively limited.

Within this scope, a substantial portion of existing work concentrates on facial expression and micro-expression analysis using event streams. Representative studies investigate facial expression recognition, action unit classification, and micro-expression analysis by exploiting the fine-grained temporal sensitivity of event cameras to capture subtle appearance changes induced by facial muscle activity~\cite{verschae2025evtransfer,cultrera2025spatio,mastropasqua2025exploring,barchid2023spiking}. To effectively model these dynamics, prior approaches employ spiking neural networks, convolutional architectures, and spatiotemporal transformers, as well as transfer learning and self-supervised representation learning strategies~\cite{barchid2023spiking,verschae2025evtransfer,cultrera2025spatio}. These works collectively demonstrate that event data is particularly well-suited for behavior-oriented facial analysis.

Another line of research explores face- and eye-related localization tasks, including face detection, eye tracking, facial landmark alignment, and pose estimation from event streams~\cite{ryan2023real,kang2025event,oral2025evaluation,iddrisu2024evaluating,kielty2025event}. These approaches typically rely on short-term motion cues generated by facial movements to estimate sparse geometric information or to support real-time tracking under varying illumination conditions~\cite{ryan2023real,kang2025event}. Multi-task learning frameworks have also been proposed to jointly perform facial landmark detection, blink detection, head pose estimation, and related tasks, aiming to improve efficiency and robustness by sharing representations across closely related facial analysis objectives~\cite{ryan2023real,kielty2025event}.

Despite these advances, existing event-based face analysis methods focus on localized or short-duration facial motions for task-specific analysis rather than identity modeling. In contrast, global facial motion, such as head rotations, can reveal richer and more stable facial structural information over time, yet remains largely unexplored in the event domain.

\subsection{Cross-Modal Adaptation for Event Data}
Learning effective representations directly from event data remains challenging, largely due to the limited scale and diversity of available annotated datasets. Compared to RGB-based vision, where large-scale data supports training deep recognition models, event-based datasets are typically smaller and task-specific, making direct training from scratch difficult to generalize~\cite{yang2023event,klenk2024masked}.

To address data scarcity, prior works have extensively explored transferring knowledge from frame-based modalities to event-based representations. A standard paradigm involves initializing event-based networks with parameters pretrained on large-scale RGB image or video datasets~\cite{hu2020learning,wu2023eventclip,chen2024velora}. By leveraging the rich visual priors learned from RGB domains, this strategy provides a robust starting point for optimization, significantly stabilizing training on smaller event datasets. Beyond simple initialization, more advanced approaches focus on explicit cross-modal alignment. For instance, self-supervised frameworks utilizing paired event-RGB data have been proposed to learn shared semantic feature spaces, allowing the strong discriminative power of RGB representations to guide the learning of event encoders~\cite{yang2023event,zhan2024spiking,liu2025leveraging}. However, these methods typically require full fine-tuning of the backbone network, which not only demands substantial computational resources but also risks overfitting or degrading the generalizability of the pretrained features when data is extremely limited.

In parallel, Low-Rank Adaptation (LoRA)~\cite{hu2022lora} has emerged as a particularly effective strategy to transfer pretrained backbones to new modalities or tasks~\cite{yang2024low}. By freezing the pretrained backbone and injecting trainable low-rank adaptation matrices into selected layers, LoRA enables efficient task-specific learning without full parameter updates. This property makes it well suited for cross-modal transfer to event data, where preserving the facial structural priors learned from RGB images is crucial to prevent overfitting and representation drift under limited training data.

\section{Dataset}
    \label{sec:dataset}
    \begin{figure}
    \centering
    \includegraphics[width=\linewidth]{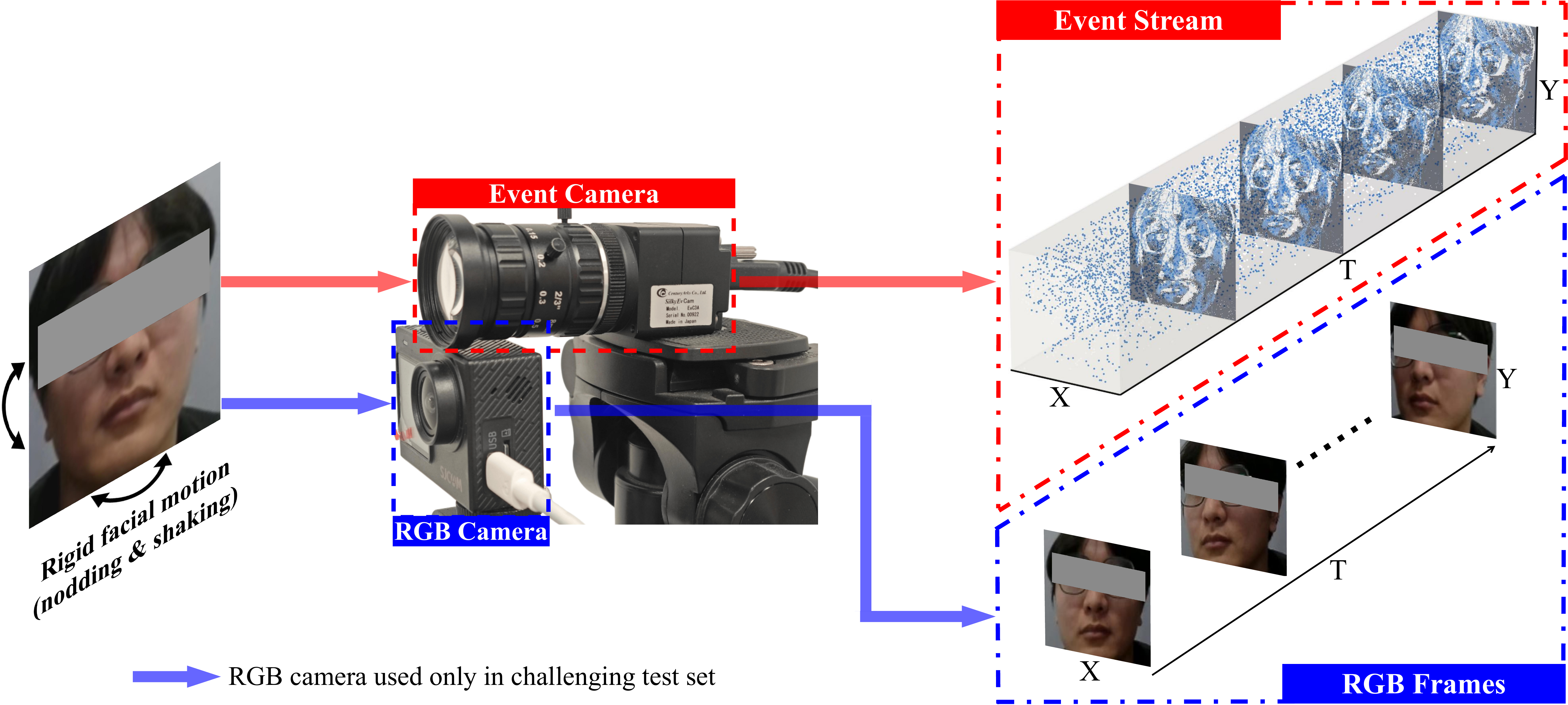}
    \caption{Overview of the EFace dataset acquisition setup. Subjects perform natural rigid facial motions (e.g., nodding and shaking) to induce intensity changes captured by the event camera, producing spatiotemporal event streams. For the challenging test subset, an RGB camera is mounted in close proximity to the event camera, and the two modalities are acquired in a synchronized manner to ensure consistent illumination conditions.}
    \label{fig:dataset_cap_pipeline}
\end{figure}

A specialized benchmark is essential for studying event-based face recognition under rigid facial motion. However, to the best of our knowledge, no public benchmark currently exists that is specifically designed for event-based face recognition driven by rigid facial motions. To fill this void, we constructed EFace, a dataset for event-based face recognition, with no identity overlap between the training and testing sets.

An overview of the acquisition setup is shown in Fig.~\ref{fig:dataset_cap_pipeline}. During data collection, the camera was positioned frontally at an approximate distance of 0.3--0.5 meters from the subject, corresponding to a near-range face acquisition setting. Since event cameras respond to relative motion or brightness variation, subjects were instructed to perform natural rigid facial motions, such as nodding and shaking, to induce stable and informative event streams while preserving facial structure. All event data were captured using a CenturyArks SilkyEvCam (Model EvC3A) equipped with a Prophesee PPS3MVCD sensor, with a spatial resolution of $640 \times 480$ and a dynamic range exceeding 120~dB.

The dataset contains 131 identities in total. The training set comprises 100 identities, collected under standard indoor and outdoor illumination using event-only acquisition. The test set contains the remaining 31 identities and is divided into two disjoint subsets. One subset, consisting of 20 identities, was collected under the same event-only acquisition settings as the training set and is used for the standard evaluation. The other subset, consisting of 11 identities, was collected under challenging low-light and side-light conditions to evaluate robustness under degraded illumination. For this subset, event streams and RGB images were captured simultaneously using a synchronized dual-camera setup with the two sensors mounted in close proximity, as illustrated in Fig.~\ref{fig:dataset_cap_pipeline}. This design was adopted only to ensure matched illumination conditions across the two modalities, thereby enabling a fair comparison between event-based and RGB-based face recognition under challenging lighting.

For preprocessing, we removed background activity and noise outside the facial region and retained only events corresponding to the face. This step suppresses irrelevant interference and makes the model focus on facial structure and motion.

\section{Methodology}
    \begin{figure*}[t]
    \centering
    \includegraphics[width=0.9\linewidth]{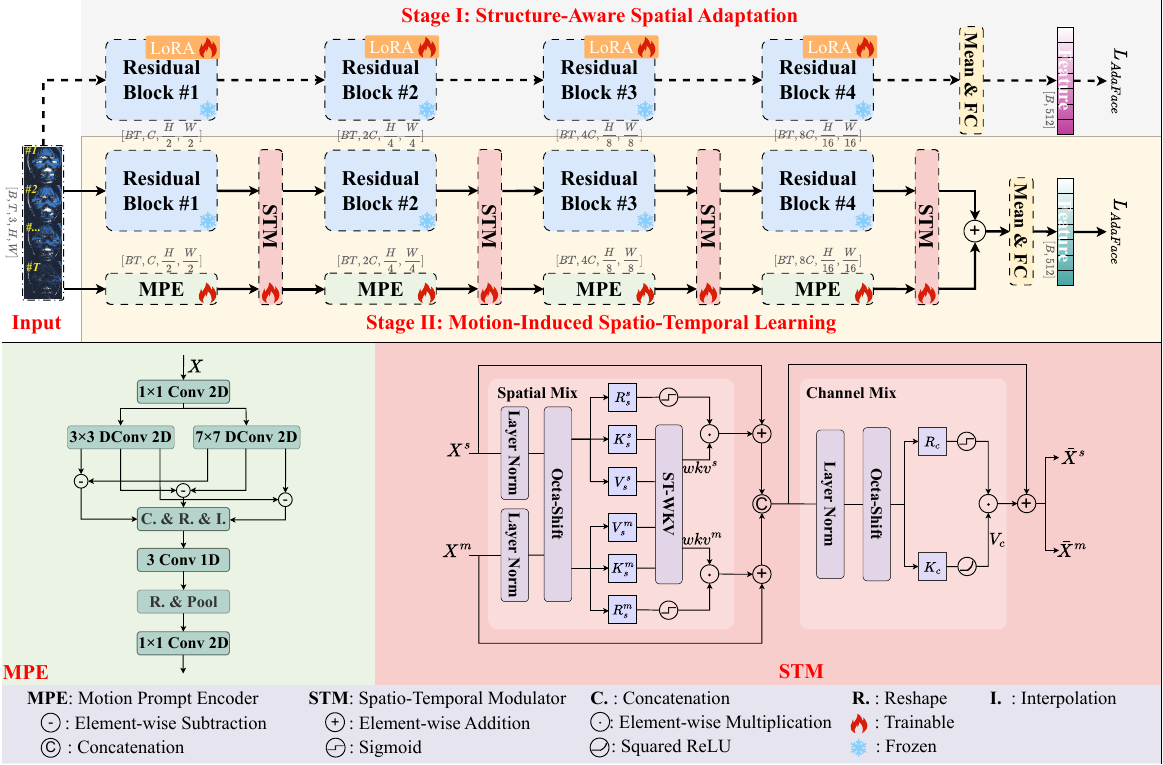}
    \caption{Overview of the proposed EventFace framework. The training process is structured into two progressive stages. In Stage I, the pretrained RGB backbone is frozen to retain structural priors, while LoRA modules are trained to align event features with the RGB domain. In Stage II, the adapted backbone is frozen. The Motion Prompt Encoder (MPE) and Spatiotemporal Modulator (STM) are introduced to work collaboratively at each stage: the MPE explicitly extracts motion prompts from event streams, which are then fused with spatial features via the STM module. This mechanism facilitates bidirectional interaction, allowing both spatial and temporal representations to be mutually refined before they are propagated to the next stage.}
    \label{fig:framework}
\end{figure*}

This section details \textbf{EventFace}, a framework designed to learn robust identity representations by synergizing static structural priors with dynamic event patterns. To overcome data scarcity, we employ a two-stage strategy that first adapts a pretrained RGB backbone to the event domain via Low-Rank Adaptation (LoRA) to transfer pretrained facial structural priors. Subsequently, we introduce the Motion Prompt Encoder (MPE) and Spatiotemporal Modulator (STM) to facilitate bidirectional interaction between motion dynamics and spatial features across network depths. The overall architecture of the proposed framework is illustrated in Fig.~\ref{fig:framework}.

\subsection{Preliminaries}
\label{subsec:preliminaries}
\subsubsection{Event Representation}
Unlike conventional frame-based cameras that capture absolute intensity frames at a fixed rate, event cameras operate asynchronously, triggered by changes in scene illumination. An event is generated at a pixel coordinate $(x, y)$ and time $t$ whenever the logarithmic intensity change exceeds a preset contrast threshold $C$. Formally, this condition is expressed as:
\begin{equation}
    \label{eq:event_generation}
    |\log(I(x, y, t)) - \log(I(x, y, t - \Delta t))| \geq C,
\end{equation}
where $I(x, y, t)$ denotes the instantaneous pixel intensity, and $\Delta t$ is the time elapsed since the last event at the same pixel. Consequently, the output of an event camera is a continuous, asynchronous stream of events $\mathcal{E} = \{e_k\}_{k=1}^{N}$, where $N$ is the total number of events. Each event $e_k$ is represented as a tuple:
\begin{equation}
    e_k = (x_k, y_k, t_k, p_k),
\end{equation}
where $(x_k, y_k)$ are the spatial coordinates, $t_k$ is the microsecond-resolution timestamp, and $p_k \in \{-1, +1\}$ indicates the polarity of the brightness change (i.e., brightness decrease or increase). This representation inherently encodes high-frequency motion dynamics and facial structural edges while filtering out redundant static background information.

\subsubsection{Input Transformation}
The raw event stream $\mathcal{E}$ is spatially sparse and temporally asynchronous, rendering it incompatible with deep CNN backbones designed for dense image data. To bridge this modality gap, we transform the continuous event stream into a sequence of dense event frames via temporal integration.

Specifically, given a fixed accumulation interval $\Delta T$, we partition the event stream into consecutive temporal windows. For the $i$-th window covering the interval $[t_{s}, t_{e})$, where $t_{e} = t_{s} + \Delta T$, we generate 2D frames by aggregating the event activities at each pixel coordinate $(x, y)$. To preserve the polarity information, we independently accumulate events for each polarity $j \in \{-1, +1\}$:
\begin{equation}
    \Gamma_{j}^{(i)}(x, y) = \sum_{e_k \in \mathcal{E}_i} \delta(x - x_k, y - y_k) \cdot \mathbb{I}(p_k = j),
\end{equation}
where $\mathcal{E}_i = \{e_k | t_{s} \leq t_k < t_{e}\}$, $\delta(\cdot)$ is the Dirac delta function, and $\mathbb{I}(\cdot)$ is the indicator function.

Subsequently, these polarity-specific density maps, denoted as $\Gamma_{+1}^{(i)}$ and $\Gamma_{-1}^{(i)}$, are rendered into a standard 3-channel representation $\mathbf{X}^{(t)} \in \mathbb{R}^{H\times W \times 3}$ using a pseudo-color mapping. This yields a synchronous input sequence $\mathcal{X} = \{\mathbf{X}^{(1)}, \mathbf{X}^{(2)}, \dots, \mathbf{X}^{(F)}\}$, where $F$ denotes the number of event frames, fully compatible with the spatial backbone. While this temporal integration inevitably compresses the microsecond-level granularity of raw events, it represents a strategic trade-off to bridge the modality gap: it sacrifices fine-grained temporal resolution to gain the dense spatial structure required for effective transfer learning. Furthermore, the reduced temporal dynamics are subsequently retrieved and modeled by the motion-sensitive design of our downstream temporal modules.

\subsection{Stage I: Structure-Aware Spatial Adaptation}
\label{subsec:stage1}

The primary objective of the first stage is to bridge the significant modality gap between RGB images and event frames while mitigating the risk of overfitting caused by limited event data. Rather than training a network from scratch, we leverage a pretrained RGB face recognition backbone to inherit its rich, generic facial structural priors.

Formally, we freeze the pretrained backbone parameters and incorporate Low-Rank Adaptation (LoRA)~\cite{hu2022lora} modules into selected target convolutional layers. Let $\mathbf{W}_0\in \mathbb{R}^{C_{out}\times C_{in}\times k \times k}$ denote the frozen weights of a target layer with input channels $C_{in}$ and output channels $C_{out}$. We approximate the weight update $\Delta \mathbf{W}$ by decomposing it into two sequential trainable components: a down-projection layer $\mathbf{W}_A$ and an up-projection layer $\mathbf{W}_B$.

For a given input feature map $\mathbf{X} \in \mathbb{R}^{H \times W \times C_{in}}$, the forward pass of the adapted layer is formulated as:
\begin{equation}
    \mathbf{H} = \mathbf{W}_0 * \mathbf{X} + \mathbf{W}_B * (\mathbf{W}_A * \mathbf{X}),
\end{equation}
where $*$ denotes the convolution operation. In this formulation, $\mathbf{W}_A\in \mathbb{R}^{r \times C_{in} \times k \times k}$ projects the input features into a low-rank latent space of dimension $r$ (where $r \ll C_{in}$), while $\mathbf{W}_B \in \mathbb{R}^{C_{out} \times r \times 1 \times 1}$ projects them back to the original output manifold. During the first training stage, the pretrained weights $\mathbf{W}_0$ remain frozen, and only the adapter parameters $\mathbf{W}_A$ and $\mathbf{W}_B$ are optimized.
Crucially, upon the completion of training, the learned residual update is analytically fused back into the original weights. This re-parameterization ensures that the inference architecture remains identical to the original backbone, without introducing additional inference-time modules. This parameter-efficient strategy allows the network to learn a modality-specific mapping that aligns the sparse event features with the RGB structural priors without the computational burden of full fine-tuning.

\begin{figure}[t]
    \centering
    \includegraphics[width=\linewidth]{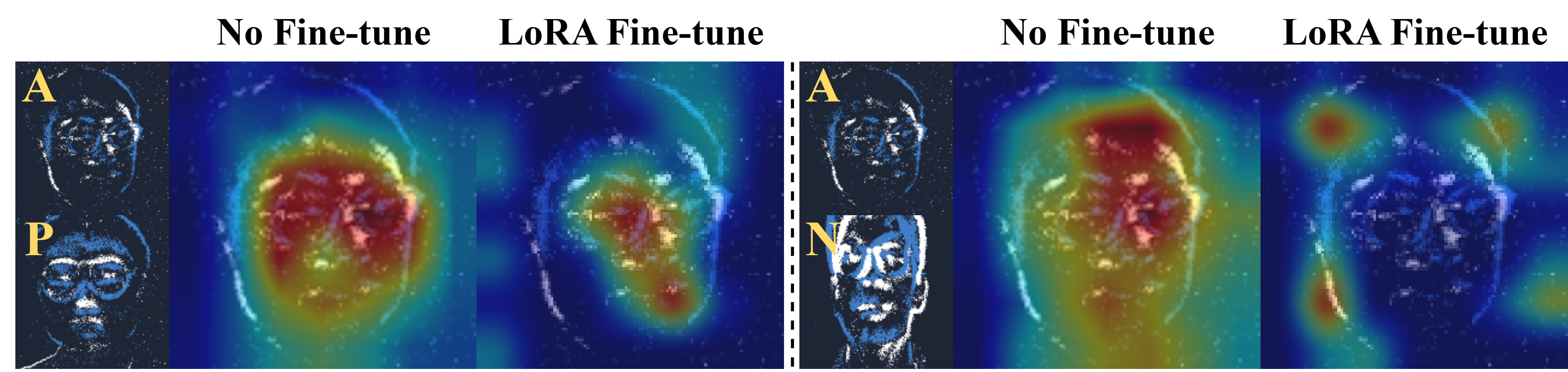}
    \caption{Spatial attention visualization via Grad-CAM. Heatmaps highlight regions that contribute most to pairwise similarity.}
    \label{fig:gradcam_lora}
\end{figure}

To empirically validate the necessity of this adaptation, we visualize the spatial attention distributions using Grad-CAM~\cite{selvaraju2017grad} in Fig.~\ref{fig:gradcam_lora}. Specifically, we compute the gradients of the cosine similarity score between the extracted feature embeddings of the input pair with respect to the final convolutional layer. This process identifies the spatial regions that contribute most significantly to the identity similarity measurement. Observing the baseline (\textit{No Fine-tune}), we find that the frozen RGB backbone can roughly localize facial regions even without adaptation, confirming the existence of shared structural priors between modalities. However, the attention patterns for same-identity (A--P) and different-identity (A--N) pairs appear nearly identical. This indicates that the frozen model only perceives the general facial shape but fails to capture the specific details needed to distinguish between different identities. In contrast, the spatial attention distribution of the \textit{LoRA Fine-tune} model concentrates on informative facial landmarks (e.g., eyes, mouth) for matched pairs, while shifting its attention significantly to the non-facial background for mismatched pairs. This suggests that Stage I improves identity-discriminative localization beyond coarse facial-structure recognition.

\subsection{Stage II: Motion-Induced Spatiotemporal Learning}
\label{subsec:stage2}
\subsubsection{Motion Prompt Encoder (MPE)}
The MPE is designed to model rigid facial motion by extracting motion-induced feature variations between consecutive frames. We conceptually interpret an event-frame sequence as comprising two factors: a relatively stable facial-structure component and a dynamic component induced by rigid head motion. Under rigid motion, the underlying semantic structures are largely preserved, while head-pose changes introduce inter-frame spatial shifts that become the dominant source of temporal variation in event features. Accordingly, the primary goal of the MPE is to extract motion-induced temporal variations as an explicit prompt for subsequent spatiotemporal modeling.

Specifically, given feature maps $\mathbf{X}_f, \mathbf{X}_{f+1} \in \mathbb{R}^{H \times W \times C}$ extracted from adjacent frames at the same semantic level, the MPE encodes motion by applying spatial aggregation at different scales. After a lightweight $1 \times 1$ convolution to reduce channel dimensionality and computational cost, we process the subsequent frame $\mathbf{X}_{f+1}$ with a larger kernel (e.g., $7 \times 7$) to obtain a more context-aware and smoothed representation. Simultaneously, the preceding frame $\mathbf{X}_f$ is processed with a smaller kernel (e.g., $3 \times 3$) to preserve localized details. The resulting features are differenced as follows:
\begin{equation}
\label{eq:mpe}
\mathbf{M}_f = \mathrm{DConv}_{7\times7}(\mathbf{X}_{f+1}) - \mathrm{DConv}_{3\times3}(\mathbf{X}_f),
\end{equation}
where $\mathrm{DConv}_{k\times k}$ denotes a depthwise convolution layer with kernel size $k\times k$, and $\mathbf{M}_f$ represents the motion feature corresponding to two adjacent frames.

In this framework, the differencing operation is employed to capture inter-frame variations associated with rigid head motion. The feature map from the preceding frame is processed with a smaller kernel to preserve local details, whereas the feature map from the subsequent frame is processed with a larger kernel to capture broader context. The resulting difference emphasizes motion-related spatial changes between consecutive frames. We compute such motion descriptors for all adjacent frame pairs in the sequence, and further apply a $1 \times 1$ convolution to aggregate these motion cues along the temporal dimension. This temporal integration helps to enhance temporal consistency and continuity of the motion representation, facilitating more reliable spatiotemporal modeling.

\begin{figure}
    \centering
    \includegraphics[width= \linewidth]{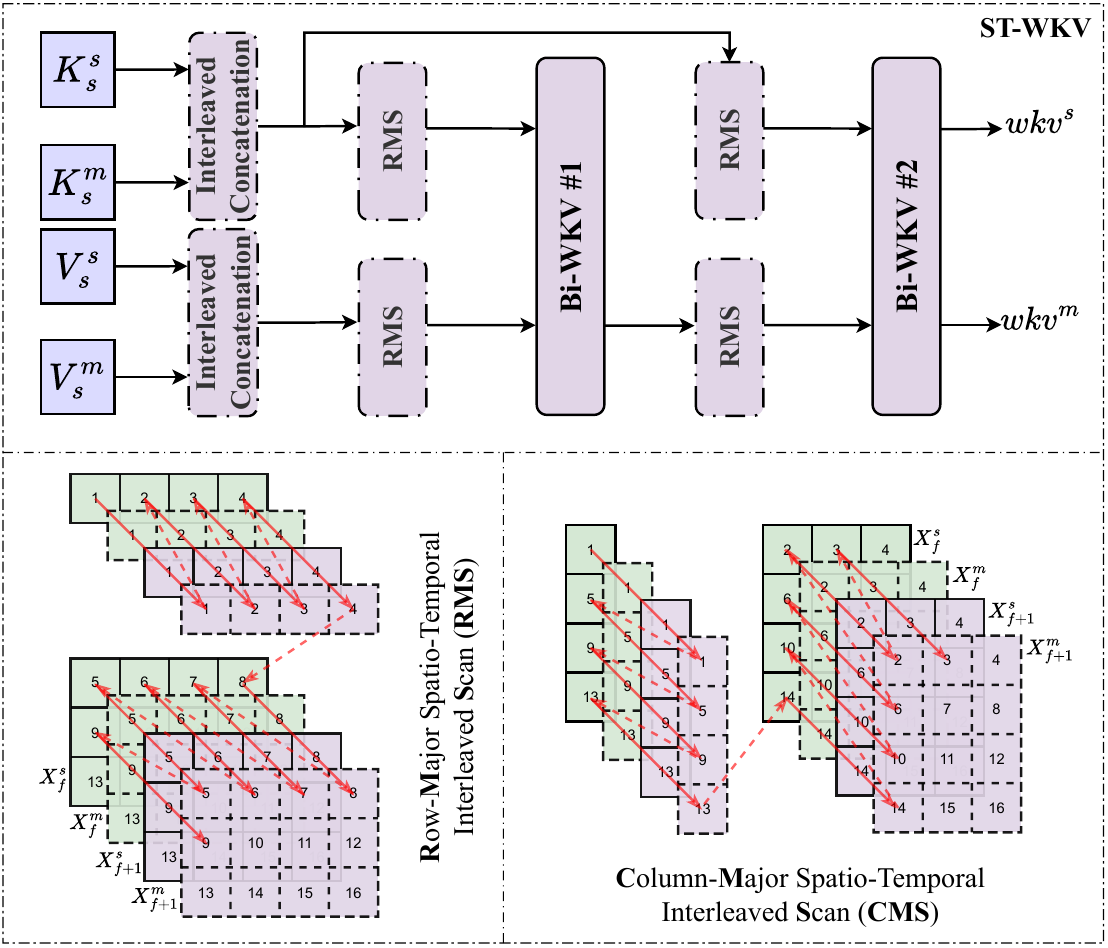}
    \caption{Illustration of the Spatiotemporal Interleaved WKV (ST-WKV). Top: The overall data flow. The module takes spatial and motion features as input, concatenates them in an interleaved manner, and processes them through cascaded Bi-WKV layers. Bottom: Visualization of the Row-Major and Column-Major Spatiotemporal Interleaved Scans.}
    \label{fig:st-wkv}
\end{figure}

\subsubsection{Spatiotemporal Modulator (STM)}
\label{subsubsec:stm}
We propose the STM to enable reciprocal modulation between the spatial features from the backbone and the motion prompts from the MPE.
Inspired by the RWKV-based visual modeling paradigm~\cite{duan2024vision,yang2025restore}, STM leverages its distance-aware global modeling capability and linear computational complexity to establish a bidirectional interaction mechanism for mutual contextualization of geometry and motion. 
To adapt this paradigm for robust spatiotemporal modeling, we introduce two core innovations within the STM architecture: the Octa-Directional Token Shift (Octa-Shift) and the Spatiotemporal Interleaved WKV Attention (ST-WKV). Specifically, Octa-Shift extends the local aggregation scope to capture omnidirectional spatial context, while ST-WKV employs an interleaved scanning strategy to enable precise global alignment between static facial structures and dynamic motion cues.

\paragraph{\textbf{Overall Architecture of STM}}
As illustrated in Fig.~\ref{fig:framework}, the Spatiotemporal Modulator (STM) adopts a two-stage design consisting of a dual-stream \emph{Spatial Mix} block followed by a unified \emph{Channel Mix} block. 
STM takes as input the spatial features from the backbone and the motion prompts from the MPE, and progressively refines them through local spatial aggregation and global spatiotemporal interaction.

\paragraph{\textbf{Spatial Mix}}
Let $\mathbf{X}^s, \mathbf{X}^m \in \mathbb{R}^{L \times C}$ denote the flattened spatial features and motion features, respectively, where $L = H \times W \times F$ represents the total sequence length derived from the spatial resolution $H \times W$ and the number of input frames $F$.
In the Spatial Mix module, the two streams are processed in parallel.
Specifically, each stream is first enhanced by a token shifting operator and then projected into three branches to obtain the receptance, key, and value representations:
\begin{equation}
\begin{aligned}
    \mathbf{R}_{sm}^\star &= \mathrm{Octa\text{-}Shift}_R(\mathrm{LN}(\mathbf{X}^\star)) \mathbf{W}_R^\star, \\
    \mathbf{K}_{sm}^\star &= \mathrm{Octa\text{-}Shift}_K(\mathrm{LN}(\mathbf{X}^\star)) \mathbf{W}_K^\star, \\
    \mathbf{V}_{sm}^\star &= \mathrm{Octa\text{-}Shift}_V(\mathrm{LN}(\mathbf{X}^\star)) \mathbf{W}_V^\star,
\end{aligned}
\end{equation}
where $\star \in \{s, m\}$ indexes the spatial stream ($s$) and the motion stream ($m$), and this notation is used consistently in the subsequent content.
Here, $\mathrm{LN}(\cdot)$ denotes Layer Normalization, and \hyperref[eq:Octa-Shift]{$\mathrm{Octa\text{-}Shift}_{(\vartriangle)}(\cdot)$} is a local token shifting operator that aggregates neighborhood information before linear projection.

The key–value pairs derived from the spatial features and the motion features are jointly processed by the 
Spatiotemporal Interleaved WKV Attention (\hyperref[para:st-wkv]{ST-WKV}) to perform a unified global aggregation that incorporates information from both feature types.
This operation produces two attention outputs,
\begin{equation}
    [\mathbf{wkv}^s, \mathbf{wkv}^m] = \mathrm{ST\text{-}WKV}(\mathbf{K}_{sm}^s, \mathbf{V}_{sm}^s, \mathbf{K}_{sm}^m, \mathbf{V}_{sm}^m),
\end{equation}
where $\mathbf{wkv}^s$ and $\mathbf{wkv}^m$ correspond to the global aggregation results for updating the spatial and motion features, respectively.

The attention outputs are then linearly projected and modulated by their corresponding receptance gates,
followed by residual connections:
\begin{align}
    \mathbf{Y}^\star &= (\sigma(\mathbf{R}_{sm}^\star) \odot \mathbf{wkv}^\star) \mathbf{W}_{O_{sm}}^\star + \mathbf{X}^\star.
\end{align}
Here, $\sigma(\cdot)$ denotes the sigmoid function (used consistently hereafter), which acts as a receptance gate to regulate the contribution of the global aggregation results. $\mathbf{W}_{O_sm}^\star$ are linear projection matrices that map the corresponding ST-WKV outputs back to the input feature space, and the residual connections preserve the original inputs
for stable optimization.

\paragraph{\textbf{Channel Mix}}
\label{channel-mix}
The outputs of Spatial Mix are then concatenated and fed into the Channel Mix block for feature integration across channels.
Formally, the two streams are combined as
\begin{equation}
    \mathbf{X}_{cm} = \mathrm{Concat}(\mathbf{Y}^s, \mathbf{Y}^m).
\end{equation}
The concatenated features are normalized and further enhanced with local spatial context,
\begin{equation}
    \mathbf{X}'_{cm} = \mathrm{Octa\text{-}Shift}(\mathrm{LN}(\mathbf{X}_{cm})).
\end{equation}
Channel-wise interactions are modeled through linear projections and a squared ReLU activation,
\begin{equation}
    \begin{split}
        \mathbf{R}_{cm}, \mathbf{K}_{cm} &= \mathbf{X}'_{cm} \mathbf{W}_R, \mathbf{X}'_{cm} \mathbf{W}_K, \\
        \mathbf{V}_{cm} &= \text{ReLU}(\mathbf{K}_{cm})^2  \mathbf{W}_V.
    \end{split}
\end{equation}
The final output of Channel Mix is obtained by gated modulation with a residual connection,
\begin{equation}
    [\bar{\mathbf{X}}^s, \bar{\mathbf{X}}^m] = (\sigma(\mathbf{R}_{cm}) \odot \mathbf{V}_{cm}) \mathbf{W}_{O_{cm}} + \mathbf{X}_{cm},
\end{equation}
which is then split back into refined spatial and motion features for subsequent processing stages.

\paragraph{\textbf{Octa-Directional Token Shift (Octa-Shift)}}
\label{octa-shift}
We introduce Octa-Shift as a token shifting operation that extends the standard horizontal and vertical shifts used in~\cite{duan2024vision} to include diagonal directions, without introducing additional learnable parameters. By incorporating diagonal neighborhood information, Octa-Shift captures local spatial relationships that do not follow purely horizontal or vertical directions, which is particularly useful for modeling facial structures and motion patterns with oblique spatial variations.

Formally, given an input token map $\mathbf{X} \in \mathbb{R}^{H \times W \times C}$, Octa-Shift augments the representation by aggregating features from eight spatial neighbors via channel-wise directional shifts. The shifted representation is defined as:
\begin{equation}
    \mathrm{Octa\text{-}Shift}_{(\vartriangle)}(\mathbf{X}) = \mathbf{X} + (1 - \mu_{(\vartriangle)}) \mathbf{X}^{\dagger},
\label{eq:Octa-Shift}
\end{equation}
where $(\vartriangle) \in \{R, K, V\}$ denotes the feature branch used in the subsequent ST-WKV module, and $\mu_{(\vartriangle)}$ is a weighting factor controlling the contribution of the shifted features.

The shifted feature map $\mathbf{X}^{\dagger}$ is constructed by first extracting a distinct channel group from each of the eight neighboring tokens around each spatial location. Specifically, for the $i$-th neighbor, the $i$-th channel group of size $C/8$ is selected. These eight channel groups, each coming from a different neighbor, are concatenated along the channel dimension to form $\mathbf{X}^{\dagger}$:
\begin{equation}
    \mathbf{X}^{\dagger} = \mathrm{Concat}\big(
    \mathcal{S}_{1}(\mathbf{X}^{(1)}_{1}),
    \mathcal{S}_{2}(\mathbf{X}^{(2)}_{2}),
    \dots,
    \mathcal{S}_{8}(\mathbf{X}^{(8)}_{8})
    \big),
\end{equation}
where $\mathbf{X}^{(i)}$ denotes the $i$-th neighboring token map of $\mathbf{X}$, $\mathbf{X}^{(i)}_i$ is its $i$-th channel group, $\mathcal{S}_i(\cdot)$ denotes the shift operation aligning the neighbor’s features to the original pixel location, and boundary tokens are handled via index clipping.

\paragraph{\textbf{Spatiotemporal Interleaved WKV Attention (ST-WKV)}}
\label{para:st-wkv}
Bi-WKV~\cite{duan2024vision} provides an efficient mechanism for global token aggregation by modeling distance-aware dependencies with linear complexity.
However, directly applying Bi-WKV to spatiotemporal modeling is suboptimal for our task, as spatial appearance features and motion cues are generated from different sources and follow distinct structural patterns. Treating them as independent sequences prevents effective alignment between static facial structures and motion-induced variations.

To address this limitation, we propose the ST-WKV, which adapts the Bi-WKV operator to jointly aggregate spatial and motion features through a structured interleaving strategy.
Given a key–value sequence $(\mathbf{K}, \mathbf{V})$, the Bi-WKV output for the $\ell$-th token is computed as
\begin{multline}
\label{eq:bi-wkv}
\mathrm{Bi\text{-}WKV}(K,V)_\ell = \\
\shoveleft{
\frac{
\sum\limits_{i=0,\, i\neq \ell}^{L-1}
\exp\!\left(
-\frac{|\ell-i|-1}{L}\, w + k_i
\right)\, v_i
+
\exp\!\left(u + k_\ell\right)\, v_\ell
}{
\sum\limits_{i=0,\, i\neq \ell}^{L-1}
\exp\!\left(
-\frac{|\ell-i|-1}{L}\, w + k_i
\right)
+
\exp\!\left(u + k_\ell\right)
}
},
\end{multline}
where $\ell$ denotes the token index in the interleaved sequence, 
$L$ denotes the sequence length, and $w$ and $u$ are learnable channel-wise parameters controlling distance decay and the current-token bonus, respectively, and $k_i$ and $v_i$ denote the $i$-th elements of the key and value sequences.

In ST-WKV, we construct a spatiotemporal token sequence by interleaving spatial tokens and motion tokens, such that each spatial location is explicitly paired with its corresponding motion cue. The interleaved sequence is then processed by the Bi-WKV operator, where the token index $\ell$ in Eq.~\eqref{eq:bi-wkv} corresponds to an interleaved spatiotemporal position. Under this formulation, the distance term $|\ell-i|$ naturally captures both spatial proximity and temporal correspondence between spatial and motion features.

To ensure comprehensive global aggregation, ST-WKV applies Bi-WKV under two complementary scanning orders: a row-major and a column-major spatiotemporal interleaved scan, as shown in Fig.~\ref{fig:st-wkv}. The outputs from the two scans are sequentially aggregated and then split back into spatial and motion components, yielding two global representations.

\subsection{Training Objective}
\label{subsec:training_objective}

Both Stage~I and Stage~II are trained using the AdaFace loss~\cite{kim2022adaface} as the unified supervision objective.
In Stage~I, AdaFace is applied to the spatially adapted features to facilitate identity-preserving domain alignment from RGB to the event domain.
In Stage~II, the same loss is imposed on the spatiotemporally enhanced representations produced by the proposed motion-aware modules, enabling robust identity discrimination under dynamic facial motion.

Using a consistent metric learning objective across both stages ensures coherent optimization of spatial adaptation and spatiotemporal modeling.

\section{Experiments}
    \subsection{Experimental Setup}
\label{subsec:setup}
\subsubsection{Evaluation Metrics and Protocols}
We evaluate our method on the proposed EFace benchmark, where the training set (100 identities) and the testing set (31 identities) share no identity overlap, as detailed in Section \ref{sec:dataset}. This setting is used to assess the model's generalization ability to unseen identities.

For the main evaluation on the EFace test set, we report both verification and identification performance. For verification, we construct all possible genuine and impostor pairs from the test set and evaluate performance in a $1\text{:}1$ matching setting. The results are summarized by the receiver operating characteristic (ROC) curve. We report the area under the ROC curve (AUC) as a threshold-independent summary metric, together with the true acceptance rate (TAR) at fixed false acceptance rates (FARs). The equal error rate (EER) is also reported.

For identification, we evaluate the model under a closed-set $1\text{:}N$ identification protocol on the test set, and report the cumulative match characteristic (CMC) curve. In particular, the Rank-1 identification rate is reported, which indicates the proportion of probe samples whose correct identity is retrieved as the top match in the gallery. Since all probe identities are included in the gallery, this evaluation is treated as closed-set identification.

In addition to the main benchmark evaluation, we perform a dedicated analysis of illumination robustness on the subset of 11 identities for which paired event and RGB data were acquired under both standard and degraded lighting conditions. In this protocol, samples captured under standard illumination serve as the gallery, while samples captured under low-light and side-light conditions serve as probes. By comparing feature distributions and cosine similarities across the two modalities under matched illumination settings, we assess their relative robustness to illumination degradation.

\subsubsection{Implementation Details} 
As detailed in Section \ref{subsec:preliminaries}, we transform the raw asynchronous events into a sequence of synchronous 3-channel frames. In our experiments, the accumulation interval is set to $\Delta T = 50$ ms. For the spatiotemporal modeling, we sample sequences of $T=4$ consecutive frames as input, covering a total temporal receptive field of 200 ms. All generated frames are resized to a spatial resolution of $112 \times 112$ pixels and normalized to $[-1, 1]$. Our framework utilizes AdaFace~\cite{kim2022adaface} with a ResNet-50 backbone initialized with weights pre-trained on the WebFace4M~\cite{zhu2021webface260m} dataset. The training relies on the two-stage optimization strategy detailed in Section \ref{subsec:stage1} and Section \ref{subsec:stage2}. In the first stage, we freeze the backbone and exclusively update the LoRA modules (rank $r=6$). Subsequently, in the second stage, the LoRA parameters are merged into the ResNet-50 backbone, after which the backbone is entirely frozen. The optimization is then restricted to the Motion Prompt Encoder (MPE) and the Spatiotemporal Modulator (STM) modules.

We optimize the network using the SGD optimizer with an initial learning rate of $5\times 10^{-5}$. The training objective is the AdaFace loss, where the margin and scale hyperparameters are fixed to $m=0.5$ and $s=32.0$ for all experiments. The batch size is fixed at 80 across both training stages, with Stage I trained for 60 epochs and Stage II for 20 epochs. 
All models were trained on an NVIDIA GeForce RTX 3090 GPU with 24 GB of memory.

\subsection{Ablation Studies}

\begin{table}[t]
\centering
\renewcommand{\arraystretch}{1.2}
\caption{Effectiveness of the two-stage framework. The backbone is AdaFace pretrained on the WebFace4M dataset.}
\begin{tabular}{cccc} 
\toprule
\multicolumn{3}{c}{Components} & \multirow{2}{*}{EER (\%)$\downarrow$} \\ 
Backbone & Stage I & Stage II & \\ 
\midrule
\checkmark & & & 28.81 \\
\checkmark & \checkmark & & 7.06 \\
\checkmark & & \checkmark & 10.57 \\
\rowcolor{gray!20}
\checkmark & \checkmark & \checkmark & \textbf{5.35} \\ 
\bottomrule
\end{tabular}
\label{abl:twostage}
\end{table}

\textit{Effectiveness of the Two-Stage Framework}: The proposed framework adopts a two-stage design to address the representation discrepancy between RGB images and event streams by jointly performing structure-aware adaptation and motion-induced modeling. To evaluate the effectiveness and necessity of each stage, a stage-wise ablation study is conducted based on a shared AdaFace~\cite{kim2022adaface} backbone pre-trained on the WebFace4M~\cite{zhu2021webface260m} dataset, as reported in Table~\ref{abl:twostage}. When both Stage I and Stage II are disabled, directly applying the RGB-pretrained backbone to the event-based benchmark results in an EER of 28.81\%, indicating a severe performance degradation caused by the modality gap. Enabling Stage I alone reduces the EER to 7.06\%, demonstrating that structure-aware adaptation effectively aligns spatial facial representations across modalities. In contrast, activating only Stage II achieves an EER of 10.57\%, suggesting that motion-induced temporal modeling provides discriminative cues but is insufficient without prior spatial adaptation. When both stages are jointly enabled, the EER is further reduced to 5.35\%, achieving the best performance among all configurations. These results indicate that Stage I and Stage II contribute complementary benefits, where spatial structure alignment establishes a robust representation foundation and motion-induced modeling further enhances identity discrimination, validating the necessity of the two-stage design.

\begin{table}[t]
\centering
\caption{Performance comparison of different tuning strategies.} 
\renewcommand{\arraystretch}{1.2}
\begin{tabular}{lcc} 
\toprule
Strategy & Rank-1 (\%)$\uparrow$ & EER (\%)$\downarrow$ \\
\midrule
Full Fine-Tuning   & 91.18 & 7.26 \\
Adapter Tuning     & 90.73 & 7.96 \\
\rowcolor{gray!20}
LoRA Tuning        & \textbf{92.09} & \textbf{6.81} \\ 
\bottomrule
\end{tabular}
\label{abl:tuning}
\end{table}

\textit{Analysis of Tuning Strategies}:
To evaluate the effectiveness of different transfer strategies, we compare the proposed LoRA-based tuning strategy with full fine-tuning and adapter tuning, where the adapter-based variant inserts a 3D convolutional bottleneck after each backbone stage, as reported in Table~\ref{abl:tuning}. Full fine-tuning achieves a Rank-1 accuracy of 91.18\%, which is inferior to LoRA tuning, suggesting that updating all backbone parameters may be less effective under limited event-domain supervision. Adapter tuning yields the lowest Rank-1 accuracy (90.73\%), indicating that introducing additional 3D bottleneck modules may perturb the pretrained spatial representations and increase the difficulty of optimization. In contrast, LoRA tuning achieves the best performance, with a Rank-1 accuracy of 92.09\% and an EER of 6.81\%. These results suggest that better preserving the facial structural priors encoded in the pretrained RGB backbone, while introducing lightweight low-rank updates, is more effective for cross-modal adaptation.

\begin{table}[t]
\centering
\caption{Joint ablation of motion encoding and spatiotemporal modeling strategies in Stage II.}
\renewcommand{\arraystretch}{1.2}
\setlength{\tabcolsep}{3pt}
\begin{tabular}{ccccc}
\toprule
MPE & \multicolumn{2}{c}{STM} & \multirow{2}{*}{Rank-1 (\%)$\uparrow$} & \multirow{2}{*}{EER (\%)$\downarrow$} \\
Kernel Size & Token Shift & Token Arrangement & & \\ 
\midrule
5-3 & Octa-Shift & Interleaved & 93.65 & 5.82 \\
7-3 & Q-Shift & Interleaved & 93.20 & 6.45 \\
7-3 & Octa-Shift & Sequential & 92.24 & 6.18 \\ 
\rowcolor{gray!20}
7-3 & Octa-Shift & Interleaved & \textbf{94.19} & \textbf{5.35} \\ 
\bottomrule
\end{tabular}
\label{abl:mpe-stm}
\end{table}

\textit{Joint Analysis of MPE and STM Designs}: 
Table~\ref{abl:mpe-stm} presents a component-wise ablation study of the Stage II architecture. All configurations are evaluated under the same Stage I setting, ensuring that the performance differences are solely attributed to Stage II design choices.
Regarding motion encoding, reducing the primary convolutional kernel in the MPE from $7\times7$ to $5\times5$, while keeping the $3\times3$ reference kernel fixed, leads to a noticeable performance drop. This suggests that a larger receptive field discrepancy is more effective in capturing spatial displacements caused by rigid facial motions. 
In terms of spatial modeling, replacing the proposed Octa-Shift with the standard Q-Shift~\cite{duan2024vision} leads to a clear degradation in accuracy, indicating that modeling spatial relationships beyond purely horizontal and vertical directions is important for capturing facial structures and motion patterns with oblique spatial variations. 
Notably, the token arrangement strategy proves to be the dominant factor for spatiotemporal fusion. 
The ``Sequential'' configuration, which concatenates spatial and motion features block-wise along the frame dimension (i.e., $[X^s_{1:F}, X^m_{1:F}]$), yields the lowest Rank-1 accuracy of 92.24\%.
This can be attributed to the exponential time-decay property inherent in the Bi-WKV~\cite{duan2024vision} operator, which prevents effective interaction between temporally distant spatial and motion blocks. 
In contrast, our proposed ``Interleaved'' strategy minimizes the temporal distance between complementary features, explicitly aligning motion cues with their corresponding structural semantics, thereby achieving the best overall performance of 94.19\%.

\begin{figure}[t]
\centering
\includegraphics[width=0.8\linewidth]{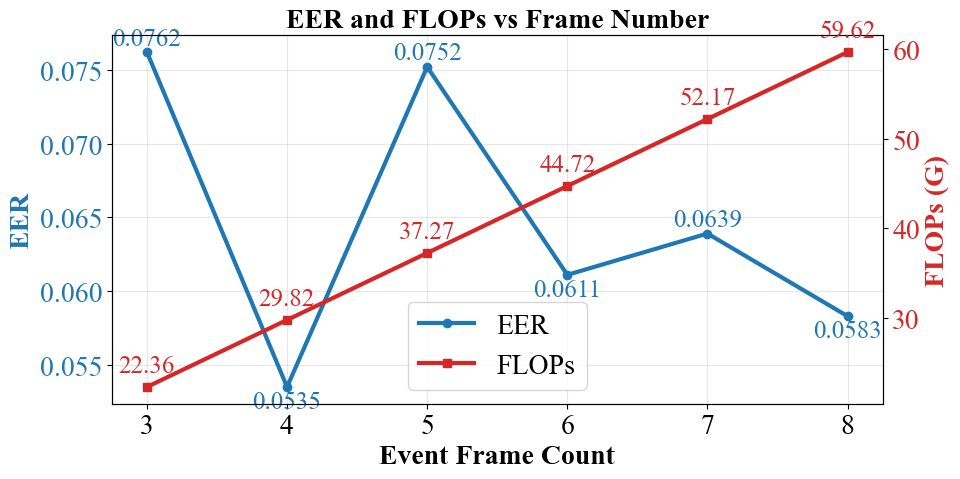} 
\caption{Effect of the input event-frame count on recognition performance and computational cost. Each event frame is accumulated over a fixed 50 ms interval.}
\label{fig:frame_ablation}
\end{figure}
\textit{Impact of the Input Frame Count}: 
We investigate the trade-off between recognition performance and computational efficiency by varying the input event frame count, as shown in Fig.~\ref{fig:frame_ablation}. 
Given that each event frame is integrated over a fixed interval of $\Delta T=50$ ms, the sequence length $T$ directly determines the total temporal receptive field (i.e., $T \times 50$ ms). 
The results indicate that $T=3$ (150 ms) yields suboptimal performance (7.62\% EER), suggesting that the temporal window is insufficient to capture sufficient spatial displacement induced by rigid facial motions. 
The optimal performance is achieved at $T=4$ (200 ms), where the model strikes the best balance between accumulating salient motion cues and minimizing the interference of non-informative frames. 
Extending the sequence beyond this point ($T > 4$) leads to performance degradation, as the prolonged duration ($> 200$ ms) introduces data redundancy without providing additional distinct identity cues, thereby complicating the optimization. 
Consequently, we adopt $T=4$ to ensure robust recognition with a manageable computational cost (29.82G FLOPs).

\subsection{Comparison with Existing Methods}

\begin{figure*}[t]
  \centering
  \captionsetup[subfloat]{font={footnotesize}}
  \subfloat[CMC\label{fig:cmc}]{
    \includegraphics[width=0.3\textwidth]{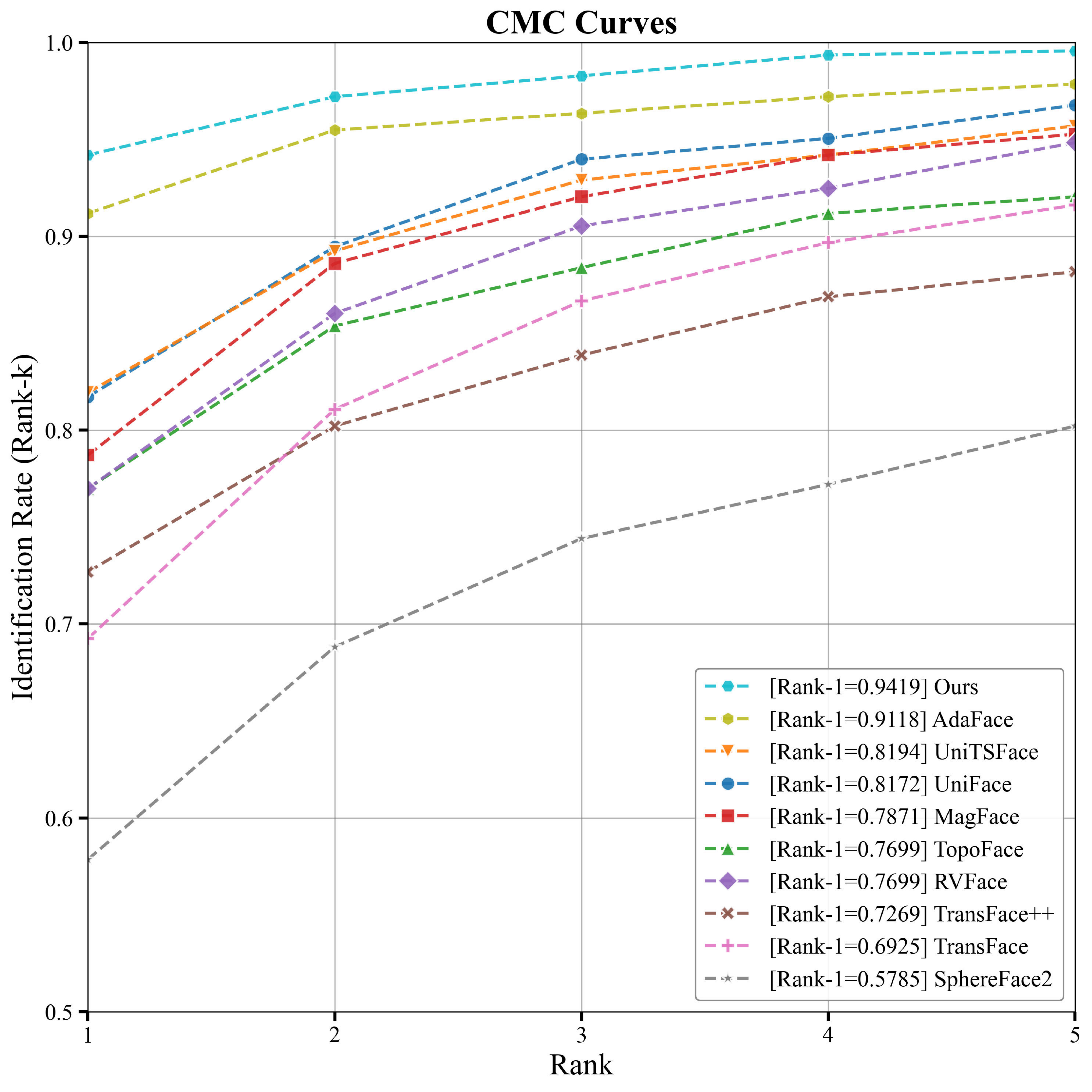}
  }
  \hfill
  \subfloat[DET\label{fig:det}]{
    \includegraphics[width=0.3\textwidth]{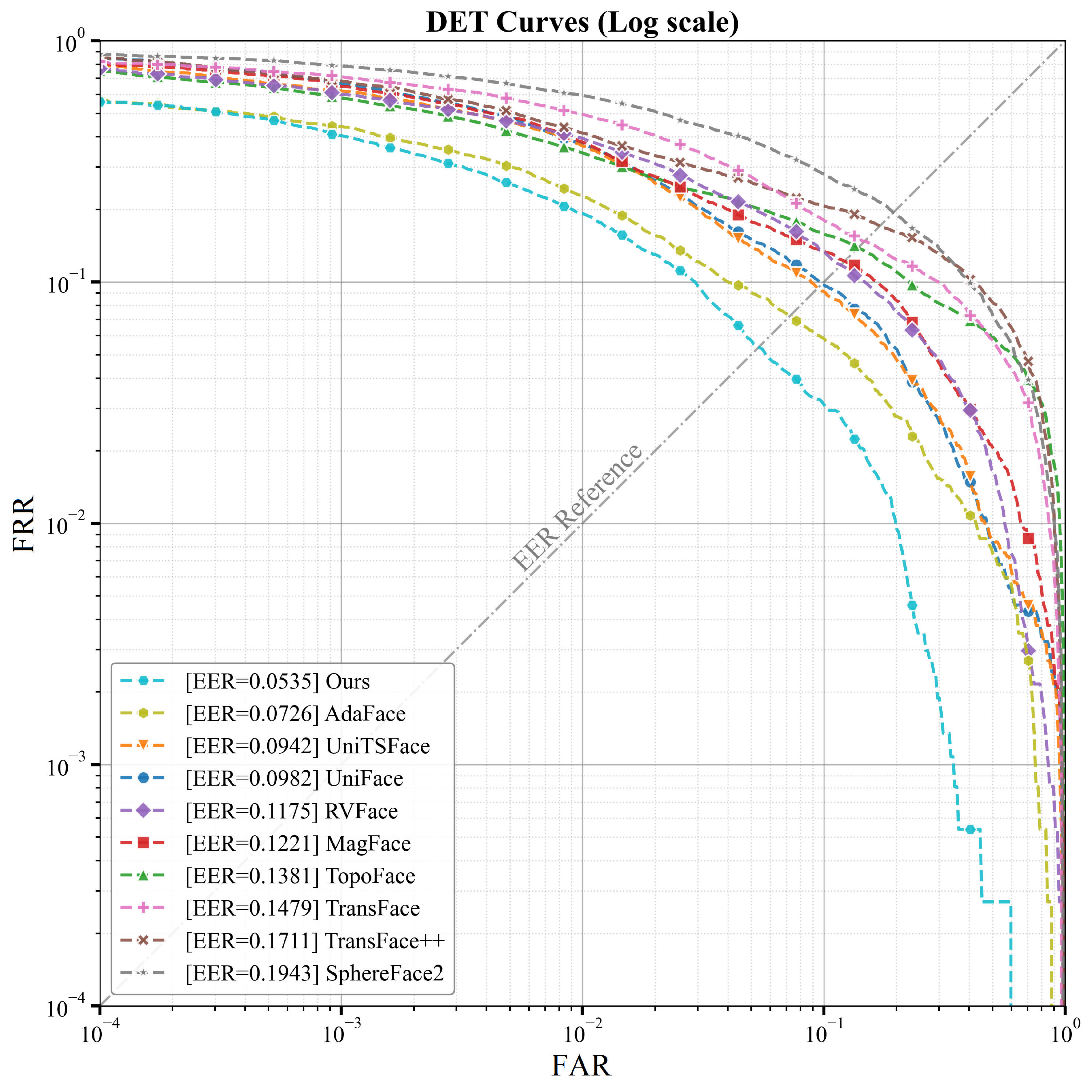}
  }
  \hfill
  \subfloat[ROC\label{fig:roc}]{
    \includegraphics[width=0.3\textwidth]{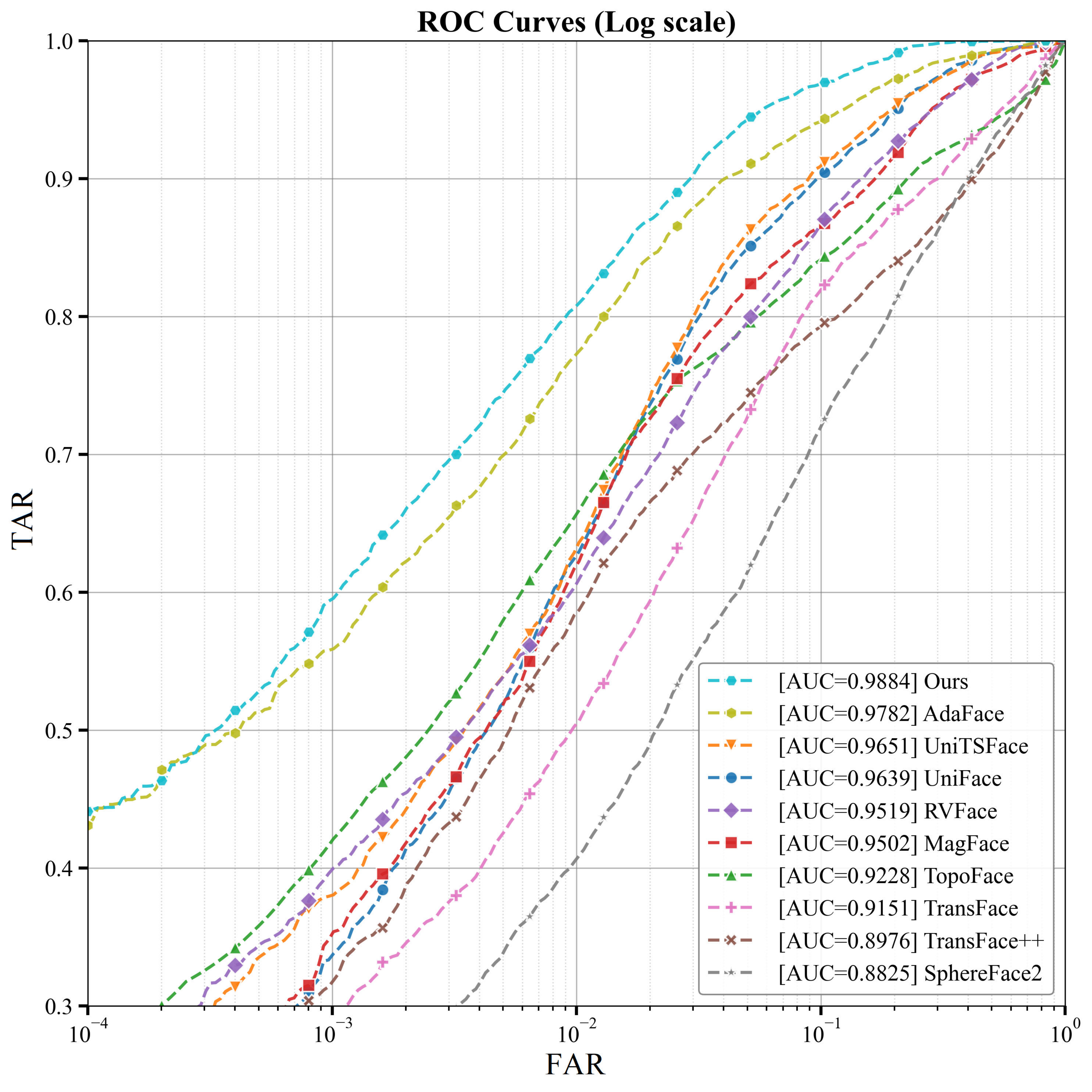}
  }
  \caption{CMC, DET, and ROC curves of EventFace and competing methods on the EFace benchmark. EventFace achieves the best overall identification and verification performance among the evaluated methods.}
  \label{fig:cmc-det-roc}
\end{figure*}

We compare the proposed method with a diverse set of representative face recognition approaches, including AdaFace~\cite{kim2022adaface}, MagFace~\cite{meng2021magface}, SphereFace2~\cite{wen2021sphereface2}, UniFace~\cite{zhou2023uniface}, UniTSFace~\cite{jia2023unitsface}, RVFace~\cite{wang2022rvface}, TopoFace~\cite{dan2024topofr}, TransFace~\cite{dan2023transface}, and TransFace++~\cite{dan2025transface++}, which represent strong and widely adopted baselines in face recognition literature.
Although all compared methods were originally developed for RGB-based face recognition, we adapt them to the event-based setting by retraining each model on the EFace training set, following a unified evaluation protocol. Specifically, all models are initialized using their officially released pretrained weights and subsequently optimized on EFace, ensuring a fair and competitive comparison.

\textit{Identification Performance Comparison}:
Fig.~\ref{fig:cmc} presents the CMC curves of all compared methods on the EFace test set. As shown, the proposed method consistently outperforms all compared baselines across the entire rank range. In particular, our approach achieves the highest Rank-1 identification rate, demonstrating a clear advantage in accurately retrieving the correct identity with minimal candidate hypotheses.

Notably, our method maintains a stable performance margin over strong baselines such as AdaFace and UniTSFace from Rank-1 to Rank-5, indicating more compact and discriminative identity embeddings. In contrast, methods primarily driven by appearance cues exhibit a larger performance gap at low ranks, suggesting reduced robustness when transferred to event-based facial representations. These results highlight the effectiveness of our structure-aware spatiotemporal modeling in capturing identity-discriminative cues under event-based sensing.

\textit{Verification Performance Comparison}:
The verification performance is further analyzed using the DET and ROC curves in Fig.~\ref{fig:det} and Fig.~\ref{fig:roc}. Overall, our method achieves the best results among the compared methods, with an equal error rate (EER) of 5.35\% and an area under the ROC curve (AUC) of 98.84\%. These metrics indicate that across the full operating range, our approach provides a favorable trade-off between false match and false non-match errors.

A closer examination of the DET curves in Fig.~\ref{fig:det} shows that EventFace maintains consistently low false non-match rates (FNMRs) across a wide range of false match rates (FMRs). In the extremely stringent regime ($\text{FMR} \approx 10^{-4}$), EventFace exhibits performance that is comparable to the re-trained AdaFace baseline. This behavior suggests that under stringent operating constraints, verification decisions are primarily governed by stable structural features. In this regime, rigid motion features mainly serve as complementary cues that reinforce identity consistency, without introducing additional noise or causing performance degradation, which is also reflected by the smooth behavior of the DET curves at very low FMRs.

As the operating point becomes less stringent and the FMR increases beyond $3 \times 10^{-4}$, motion information begins to play a more prominent role. By alleviating identity ambiguities that remain when relying solely on structural features, motion-aware modeling enables EventFace to achieve a more favorable balance between false acceptance and false rejection, resulting in consistently lower FNMRs than all compared methods.

The ROC curves in Fig.~\ref{fig:roc} further support these observations. EventFace achieves the largest AUC and maintains higher true acceptance rates across most operating points, demonstrating that the proposed method delivers strong and stable verification performance over a broad range of thresholds rather than being optimized for a single operating point.

\subsection{Robustness to Degraded Illumination}
A key premise of this study is that the High Dynamic Range (HDR) nature of event cameras confers intrinsic robustness against illumination degradation, a scenario where conventional frame-based systems typically degrade substantially. To empirically validate this, we conduct a comparative analysis using the Comparative Robustness Protocol defined in Section~\ref{subsec:setup}, where the Gallery is established using samples captured under standard indoor illumination, while the Probe comprises data subjected to extreme low-light (Levels 1–3) and sidelight conditions.

\begin{figure}[t]
  \centering
  \captionsetup[subfloat]{font={footnotesize}}
  \subfloat[Event modality\label{fig:event_sim}]{
    \includegraphics[width=0.5\linewidth]{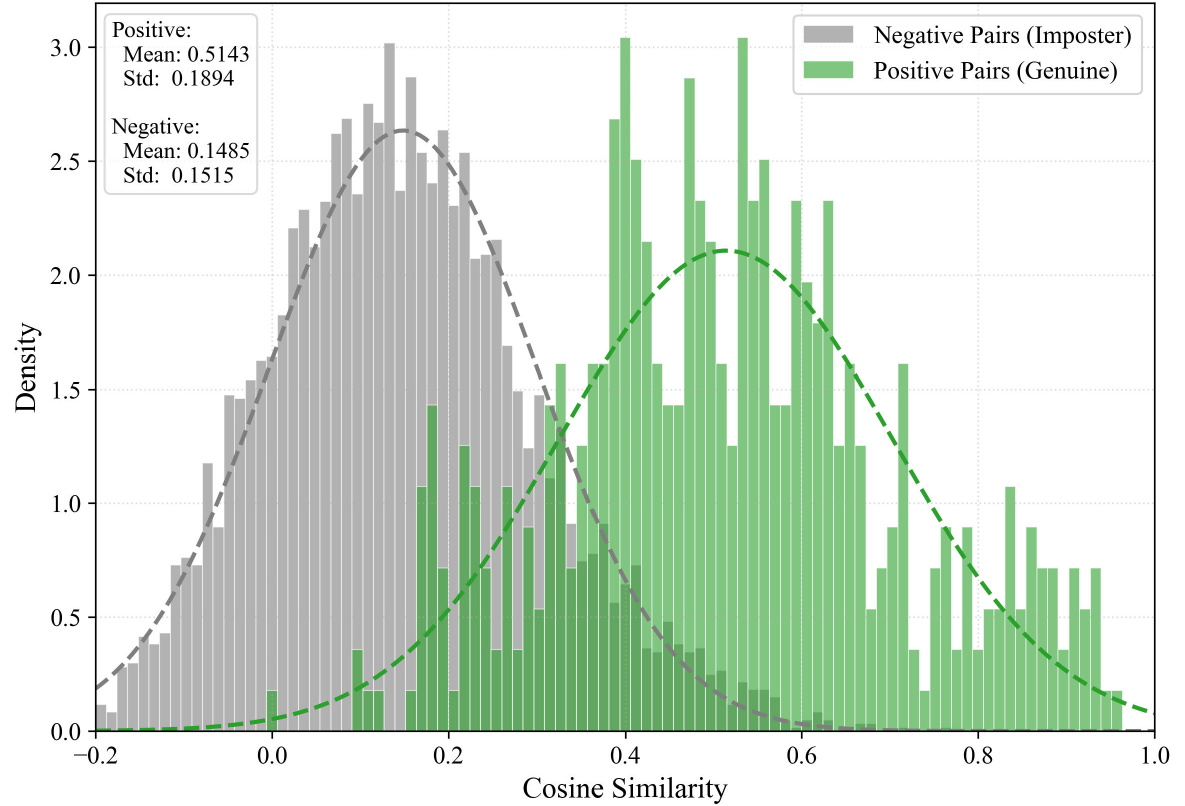}
  } 
  \subfloat[RGB modality\label{fig:rgb_sim}]{
    \includegraphics[width=0.5\linewidth]{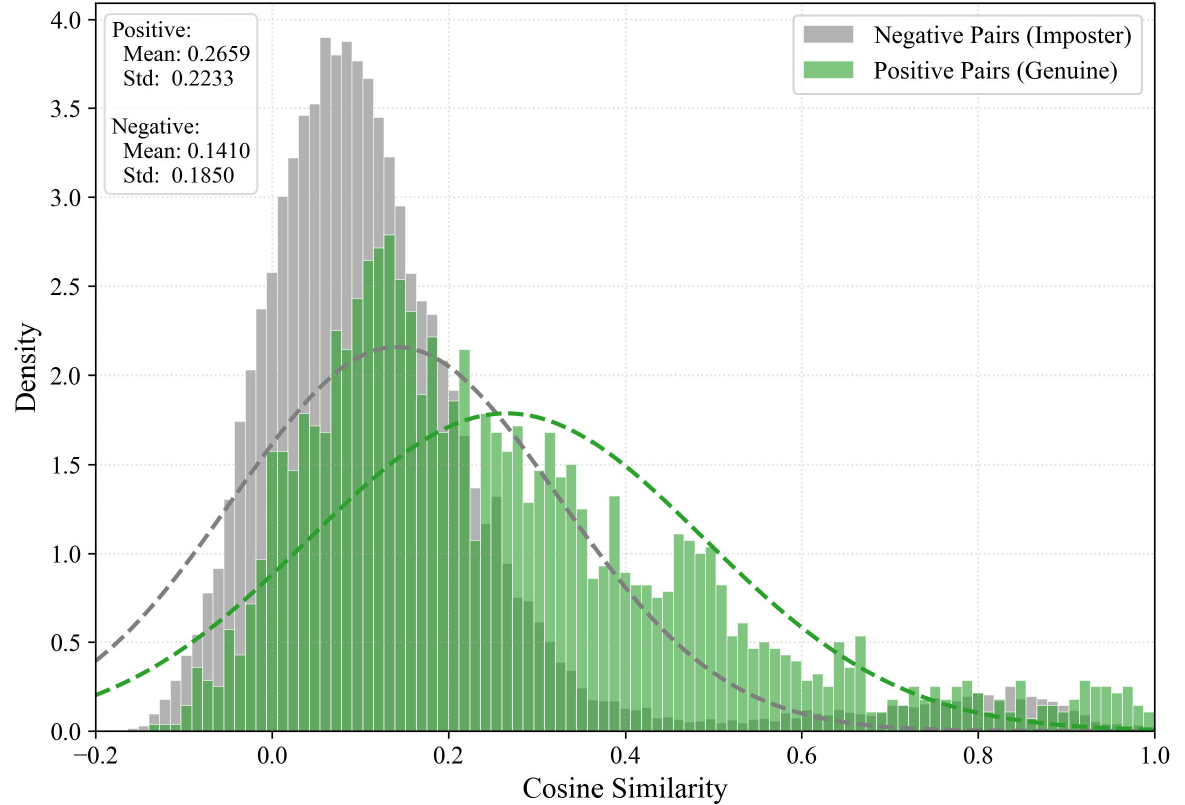}
  }
  
  \caption{Cosine similarity distributions under the degraded illumination challenge. (a) Event modality and (b) RGB modality. The plots illustrate the distributions of genuine (green) and impostor (gray) matching pairs.}
  \label{fig:modalities_sim}
\end{figure}

\textit{Feature Space Distribution Analysis}:
We first investigate the discriminability of the learned embeddings by visualizing the cosine similarity distributions for both genuine (positive) and impostor (negative) pairs. Fig.~\ref{fig:event_sim} and Fig.~\ref{fig:rgb_sim} illustrate the probability density functions of the matching scores for event and RGB modalities, respectively. As shown in Fig.~\ref{fig:rgb_sim}, the RGB modality suffers from severe feature space collapse under degraded illumination. Due to the limited dynamic range of standard sensors, the facial structure in the Probe images is obscured by noise and underexposure. Consequently, the distribution of positive pairs (green) shifts significantly toward the left, heavily overlapping with the negative distribution (grey). The small margin between the mean scores of positive ($\mu=0.2659$) and negative pairs ($\mu=0.1410$) indicates that the model struggles to distinguish genuine identities from impostors, leading to a high false rejection rate.In stark contrast, the event-based embeddings in Fig.~\ref{fig:event_sim} exhibit remarkable stability. Despite the significant domain gap between the standard-light Gallery and the degraded-light Probe, the positive distribution remains well-separated from the negative distribution. The mean similarity of positive pairs ($\mu=0.5143$) is substantially higher than that of the negative pairs ($\mu=0.1485$), maintaining a clear decision boundary. This statistical evidence confirms that our structure-driven spatiotemporal features are invariant to photometric variations, effectively preserving identity discriminability even when the visual appearance is compromised.

\begin{figure}
    \centering
    \includegraphics[width=0.7\linewidth]{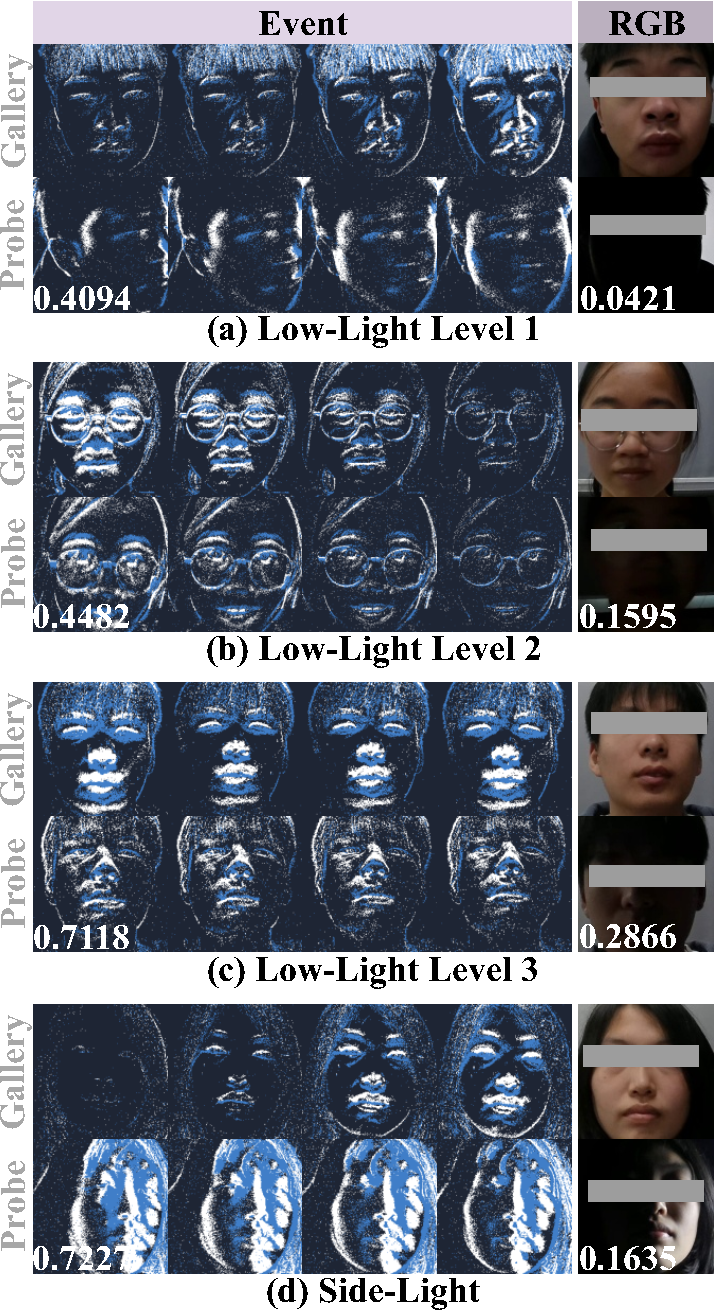}
    \caption{Visualization of matching examples under the degraded illumination challenge. The Gallery samples are captured under standard indoor lighting, while the Probe samples correspond to varying low-light levels (Levels 1–3) and sidelight conditions. The computed cosine similarity scores are annotated for both RGB and Event modalities.}
    \label{fig:degraded_vis}
\end{figure}

\textit{Qualitative Case Study}:
Fig.~\ref{fig:degraded_vis} visualizes specific matching examples under varying illumination challenges, accompanied by their corresponding cosine similarity scores. Crucially, all Gallery samples are captured under standard indoor lighting to serve as references, while the Probe samples are subjected to significant environmental degradation. We categorize the low-light scenarios into three escalating levels of difficulty: Level 3 represents a dim environment where the face is visible but suffers from severe detail loss, rendering facial features hard to distinguish; Level 2 introduces further degradation; Level 1 corresponds to a condition of near-total darkness. As illumination drops across these levels, the RGB inputs rapidly deteriorate, eventually becoming nearly indistinguishable black images in Level 1. Consequently, the matching confidence for RGB plummets to near-random levels (e.g., dropping to 0.0421 in Level 1), indicating a complete failure of photometric feature extraction against the normal-light gallery. 
In contrast, the event-based representations preserve clear facial contours and salient structural landmarks, such as the eyes and mouth. Even under the pitch-dark Level~1 condition, the event stream yields a stable similarity score (e.g., 0.4094), demonstrating its ability to capture structural information that is inaccessible to conventional frame-based sensors.
In the sidelight scenario, due to strong illumination asymmetry, the RGB image exhibits severe photometric imbalance, which adversely affects recognition performance, dropping the similarity score to 0.1635. By contrast, since event cameras respond to relative intensity changes rather than absolute brightness, the resulting event stream preserves a high degree of spatial consistency across the face. This leads to a substantially higher matching confidence, reaching a score of 0.7227.

\subsection{Privacy Analysis}
\begin{figure}
    \centering
    \includegraphics[width=0.8\linewidth]{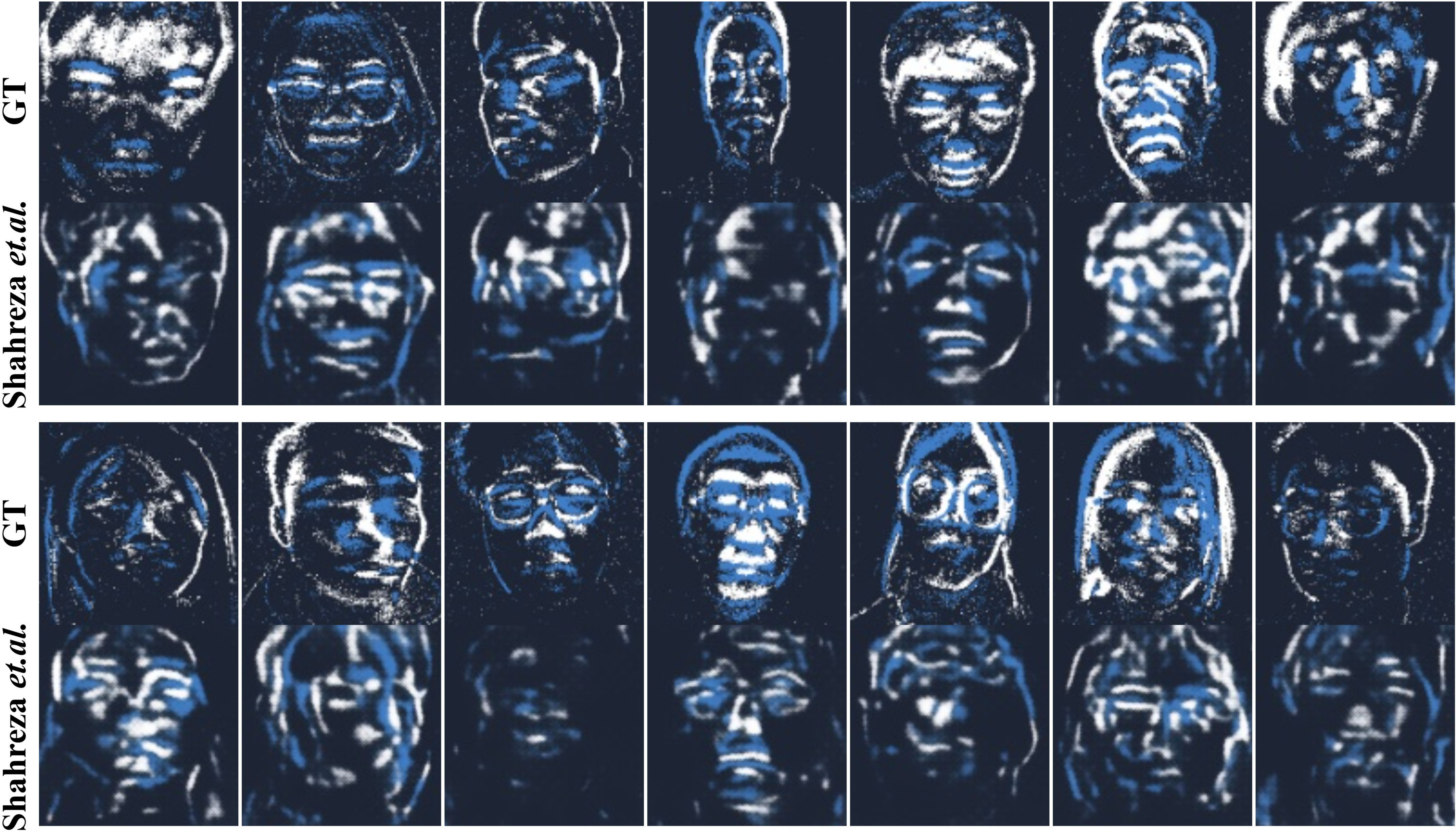}
   \caption{Visualization of face reconstruction from leaked feature templates. The top rows (GT) display the ground-truth Event Frames, while the bottom rows (Shahreza \textit{et al.}) show the reconstructed results.}
    \label{fig:temp_const}
\end{figure}

\begin{figure}
    \centering
    \includegraphics[width=0.8\linewidth]{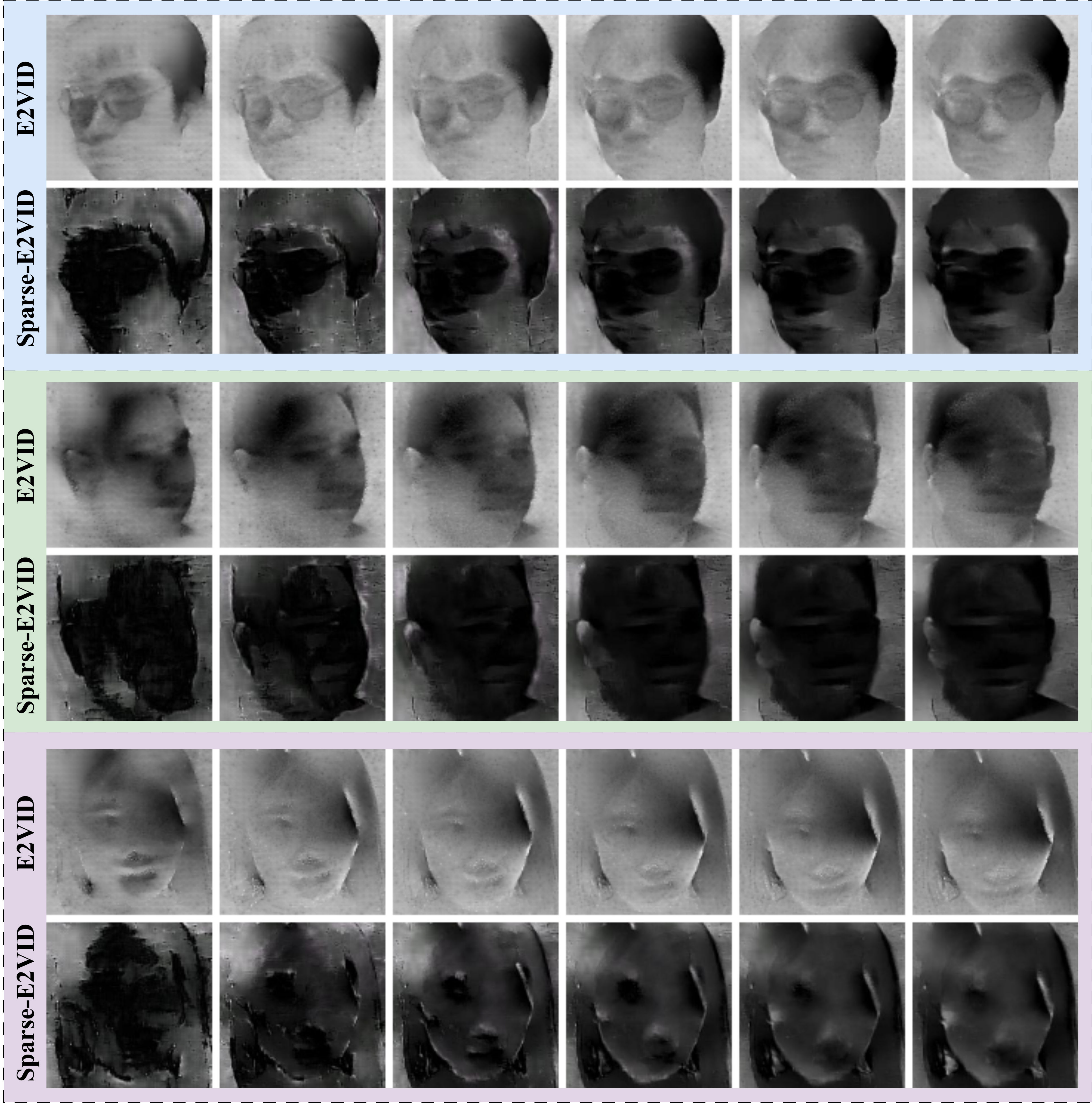}
    \caption{Privacy evaluation under a worst-case raw data leakage scenario. We employ representative event-to-video reconstruction algorithms, E2VID and Sparse-E2VID, to recover visual information directly from the 200 ms raw event streams.}
    \label{fig:eventstream_constr}
\end{figure}

\textit{Template Reconstruction Analysis}:
To evaluate the privacy-related characteristics of EventFace representations, we adapt a face template reconstruction method originally proposed by~\cite{shahreza2024face}. for the RGB domain to the event-based setting. Under this experimental setting, the reconstruction model attempts to generate corresponding facial representations given the event-based face feature templates.
To ensure a fair and competitive evaluation, the reconstruction model is retrained on the EFace training set, where it is supervised to map the extracted feature templates back to the corresponding Event Frames. Considering the inherent sparsity of event data and the substantial representational gap between event-based and conventional image modalities, we perform extensive hyperparameter tuning and recalibrate the loss function weights to facilitate stable convergence under the event modality. This optimization process is intended to avoid degenerate reconstruction results caused by training instability or inappropriate configurations.

As illustrated in Fig.~\ref{fig:temp_const}, even under carefully tuned training settings, the reconstructed outputs remain highly abstract, exhibiting only weak, face-like appearances at a coarse semantic level in a very limited number of cases. The reconstructed results lack stable global geometry, consistent facial component layouts, and coherent spatial relationships, and show no structural consistency with the corresponding ground-truth faces. Since these outputs cannot be regarded as valid recoveries of the original facial structure, they do not visually support reliable recovery of fine-grained facial appearance or clearly interpretable semantic attributes.

Further analysis suggests that the limited effectiveness of template inversion in the event domain may not only stem from the sparsity and asynchronous nature of event data, which inherently complicates template reconstruction, but may also be related to the spatiotemporal deep interaction modeling adopted in EventFace. Unlike conventional RGB representations that primarily rely on static spatial features, the representations learned by EventFace result from multi-layer deep interactions between spatial structure and motion-induced temporal features. Under this mechanism, spatial features are continuously modulated by motion features at different network stages. These motion features mainly characterize relative displacements and temporal variations, and do not necessarily carry privacy-sensitive information directly useful for identity reconstruction. Such spatiotemporal interactions may reshape the original spatial patterns at the representation level, making the final templates closer to discriminative representations that incorporate dynamic contextual information. Consequently, the reversibility of the learned templates is substantially reduced.

\textit{Privacy Evaluation on Raw Event Streams}:
We further consider a stronger input leakage setting where an adversary is assumed to have direct access to the raw event streams (in 200 ms temporal windows) used as model inputs. Under this condition, the adversary can bypass the feature template inversion process and directly employ existing event-to-video reconstruction algorithms to convert the event streams into visual intensity maps. Two representative reconstruction methods were selected to process the leaked event streams~\cite{rebecq2019high,cadena2023sparse}, with the results illustrated in Fig.~\ref{fig:eventstream_constr}. From these experimental results, it can be observed that the reconstructed outputs exhibit a certain degree of facial contours and pose cues at the visualization level. Particularly in some samples, the general orientation of the head and the global facial shape are vaguely discernible, indicating that the raw event streams indeed contain certain low-level structural information that can be partially recovered by current reconstruction algorithms. However, compared to natural images or high-quality grayscale faces, these reconstructions still suffer from significant noise interference, unstable contrast, and structural blurring. The spatial relationships of critical facial regions (e.g., eyes, nose, and mouth) lack consistency, and local structures are highly susceptible to the inherent sparsity and cumulative noise of event data, making it difficult to form stable and clear facial geometry. 

Such instability is prevalent across different identities and temporal segments. Considering the absence of critical fine-grained identity markers including precise facial proportions and textural details, performing reliable identity verification using these reconstructed results may pose significant challenges. This experimental finding, to some extent, reflects the potential limitations of visual information recovery within this sensing paradigm: even in an extreme setting of raw data leakage, the inherent sparse and asynchronous sensing characteristics of event cameras may still provide a preliminary physical barrier for privacy protection. This provides a reference perspective for understanding the security of EventFace at both the perception and representation levels.

\section{Conclusion}
    \label{conclusion}
    This paper presented EventFace, a framework for event-based face recognition based on structure-driven spatiotemporal modeling. Instead of relying on stable photometric appearance as in conventional RGB-based face recognition, our method models identity from the interaction between facial structure and motion-induced event responses. To this end, we transferred structural facial priors from pretrained RGB face models to the event domain through LoRA-based adaptation, and further introduced temporal modeling modules to integrate motion cues with spatial features.

Experiments on the proposed EFace benchmark showed that EventFace achieves strong performance in both identification and verification. The results further suggest that event-based face recognition can provide improved robustness under degraded illumination, while exhibiting reduced reconstructability from leaked templates. These findings suggest that event cameras are a promising alternative sensing modality for face recognition, particularly when robustness to illumination variation and reduced visual reconstructability are desired.

At the same time, this study remains limited by the scale and acquisition diversity of the current dataset. Future work will focus on larger and more diverse event-based face datasets, broader cross-device and cross-condition evaluation, and more rigorous analysis of privacy-related properties in event-based identity representations.

\bibliographystyle{IEEEtran}
\bibliography{IEEEabrv,ref}

\begin{IEEEbiography}
[{\includegraphics[width=1in,height=1.25in,clip,keepaspectratio]{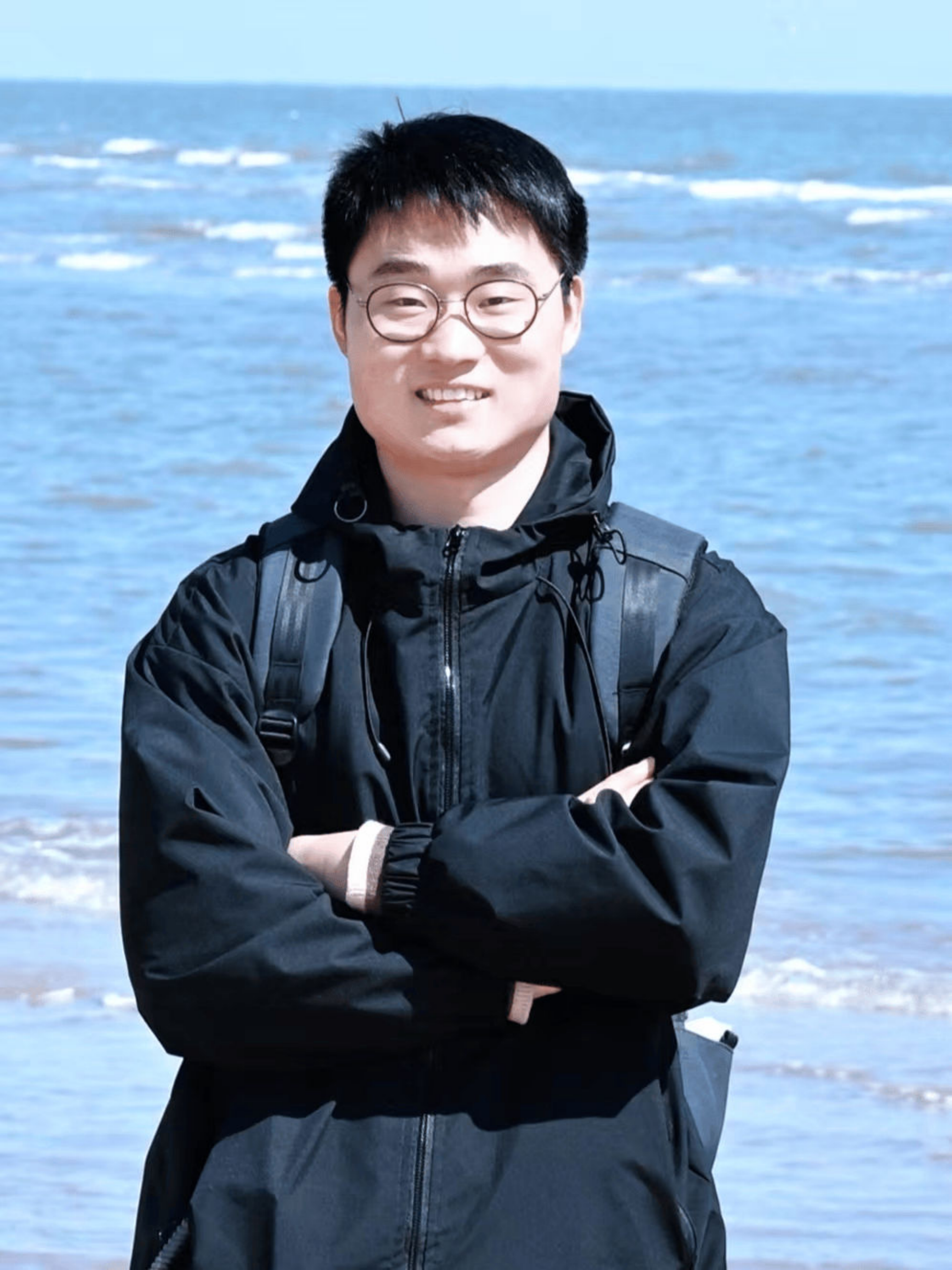}}]
{Qingguo Meng} received his B.Eng. degree in computer science and technology from Henan Polytechnic University, Jiaozuo, China, in 2022. He is currently working on his Ph.D. at the School of Artificial Intelligence, Anhui University in Hefei, China. His research directions are object tracking, low-light image enhancement, medical imaging, and biometrics.
\end{IEEEbiography}

\begin{IEEEbiography}
[{\includegraphics[width=1in,height=1.25in,clip,keepaspectratio]{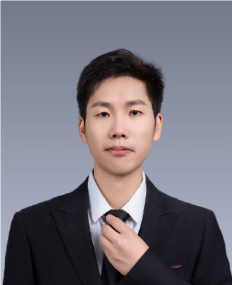}}]
{Xingbo Dong} (Member, IEEE) received the B.S. degree from Huazhong Agriculture University, Wuhan, China, in 2014, and the Ph.D. degree from the Faculty of Information Technology, Monash University, Melbourne, VIC, Australia, in 2021.
He held a post-doctoral position with Yonsei University, Seoul, South Korea, in 2022. He is currently a Lecturer with Anhui University, Hefei, China. His research interests include biometrics, medical imaging, and image processing.
\end{IEEEbiography}

\begin{IEEEbiography}[{\includegraphics[width=1in,height=1.25in,clip,keepaspectratio]{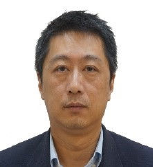}}]{Zhe Jin}
(Member, IEEE) obtained a Ph.D. in Engineering from Universiti Tunku Abdul Rahman Malaysia (UTAR). He is a Professor at the School of Artificial Intelligence, Anhui University, China. His research interests include Biometrics, Pattern Recognition, Computer Vision, and Multimedia Security. He has published over 70 refereed journals and conference articles, including IEEE Trans. IFS, SMC-S, DSC, PR. He was awarded the Marie Skłodowska-Curie Research Exchange Fellowship. He visited the University of Salzburg, Austria, and the University of Sassari, Italy, respectively, as a visiting scholar under the EU Project IDENTITY 690907.
\end{IEEEbiography}

\begin{IEEEbiography}[{\includegraphics[width=1in,height=1.25in,clip,keepaspectratio]{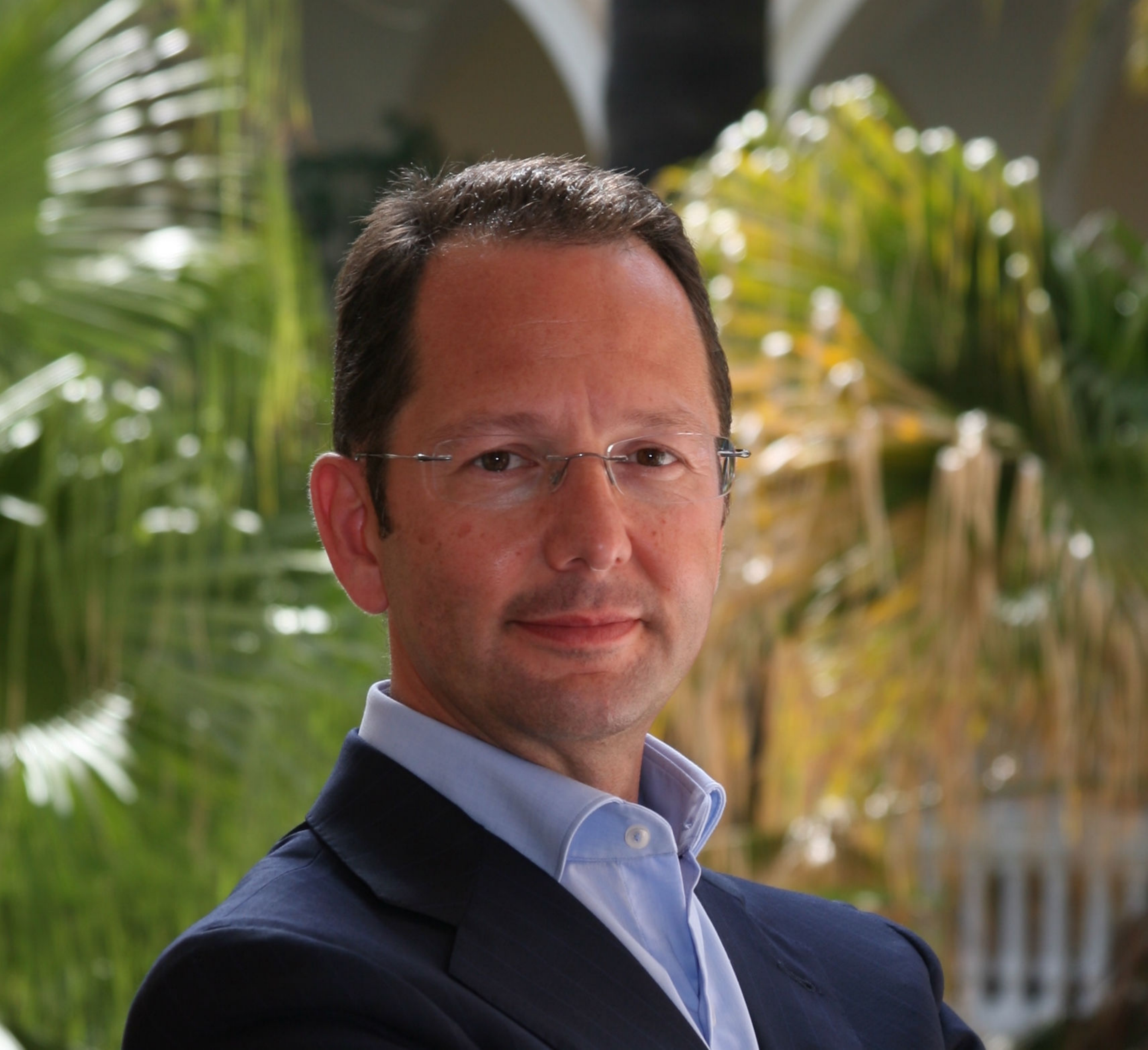}}]{Massimo Tistarelli} (Senior Member, IEEE) received the Ph.D. degree in computer science and robotics from the University of Genoa in 1991., Since 1994, he has been the Director of the Computer Vision Laboratory, Department of Communication, Computer and Systems Science, University of Genoa. He is with the University of Sassari, Italy, leading several national and European projects on computer vision applications and image-based biometrics. Since 1986, he has been involved as the project coordinator and the task manager in several projects on computer vision and biometrics funded by the European Community. Since 2003, he has been the Founding Director of the Int.l Summer School on Biometrics (now at the 16th edition: https://biometrics.uniss.it). He is currently a Full Professor in computer science (with tenure) and the Director of the Computer Vision Laboratory, University of Sassari. He is the coauthor of more than 150 scientific papers in peer-reviewed books, conferences, and international journals. His main research interests include biological and artificial vision (particularly in the area of recognition, 3-D reconstruction, and dynamic scene analysis), pattern recognition, biometrics, visual sensors, robotic navigation, and visuo-motor coordination. He is one of the world-recognized leading researchers in the area of biometrics, especially in the field of face recognition and multimodal fusion. He is a Founding Member of the Biosecure Foundation, which includes all major European research centers working in biometrics. He is a Fellow Member of the IAPR and the Vice President of the IEEE Biometrics Council. He organized and chaired several world-recognized scientific events and conferences in the area of computer vision and biometrics, and he is an Associate Editor for several scientific journals, including IEEE Transactions on Pattern Analysis and Machine Intelligence, IET Biometrics, Image and Vision Computing, and Pattern Recognition Letters. He is the Principal Editor for the Springer books Handbook of Remote Biometrics and Handbook of Biometrics for Forensic Science.
\end{IEEEbiography}

\end{document}